\documentclass[journal]{vgtc}                     

\graphicspath{{figures/}{pictures/}{images/}{./}} 

\usepackage{microtype}                 
\PassOptionsToPackage{warn}{textcomp}  
\usepackage{textcomp}                  
\usepackage{mathptmx}                  
\usepackage{times}                     
\usepackage{cite}                      
\usepackage{tabu}                      
\usepackage{booktabs}                  
\usepackage{amsmath}
\usepackage{amssymb}

\usepackage{multirow}
\newcolumntype{P}[1]{>{\centering\arraybackslash}m{#1}}
\usepackage{adjustbox}
\usepackage{subfig}

\usepackage{soul}

\setstcolor{red}

\onlineid{1221}

\vgtccategory{Research}

\vgtcpapertype{algorithm/technique}

\title{NeRF-NQA: No-Reference Quality Assessment for Scenes Generated by NeRF and Neural View Synthesis Methods}

\author{%
  \authororcid{Qiang Qu}{0000-0002-6648-5050},
  Hanxue Liang,
  \authororcid{Xiaoming Chen}{0000-0002-7503-3021}\textsuperscript{\rm *},
  \authororcid{Yuk Ying Chung}{000-0002-3158-9650}, and 
  \authororcid{Yiran Shen}{0000-0003-1385-1480}\textsuperscript{\rm *}
}

\authorfooter{
  \item
  	Qiang Qu and Yuk Ying Chung are with the School of Computer Science, the University of Sydney, Australia.
  	E-mail: \{vincent.qu, vera.chung\}@sydney.edu.au.
  \item
  	Hanxue Liang is with the Department of Computer Science and Technology, the University of Cambridge, United Kingdom.
  	E-mail: hl589@cam.ac.uk.
  \item Xiaoming Chen is with the School of Computer and Artificial Intelligence, Beijing Technology and Business University, China.
  	E-mail: xiaoming.chen@btbu.edu.cn.
   \item Yiran Shen is with the School of Software, Shandong University, China.
  	E-mail: yiran.shen@sdu.edu.

   \item Xiaoming Chen\textsuperscript{\rm *} and Yiran Shen\textsuperscript{\rm *} are the corresponding authors

}

\abstract{%
  Neural View Synthesis (NVS) has demonstrated efficacy in generating high-fidelity dense viewpoint videos using a image set with sparse views. However, existing quality assessment methods like PSNR, SSIM, and LPIPS are not tailored for the scenes with dense viewpoints synthesized by NVS and NeRF variants, thus, they often fall short in capturing the perceptual quality, including spatial and angular aspects of NVS-synthesized scenes. Furthermore, the lack of dense ground truth views makes the full reference quality assessment on NVS-synthesized scenes challenging. For instance, datasets such as LLFF provide only sparse images, insufficient for complete full-reference assessments. To address the issues above, we propose NeRF-NQA, the first no-reference quality assessment method for densely-observed scenes synthesized from the NVS and NeRF variants. NeRF-NQA employs a joint quality assessment strategy, integrating both viewwise and pointwise approaches, to evaluate the quality of NVS-generated scenes. The viewwise approach assesses the spatial quality of each individual synthesized view and the overall inter-views consistency, while the pointwise approach focuses on the angular qualities of scene surface points and their compound inter-point quality. Extensive evaluations are conducted to compare NeRF-NQA with 23 mainstream visual quality assessment methods (from fields of image, video, and light-field assessment). The results demonstrate NeRF-NQA outperforms the existing assessment methods significantly and it shows substantial superiority on assessing NVS-synthesized scenes without references.
  An implementation of this paper are available at \url{https://github.com/VincentQQu/NeRF-NQA}.
}


\teaser{
  \centering
  \includegraphics[width=1.0\textwidth]{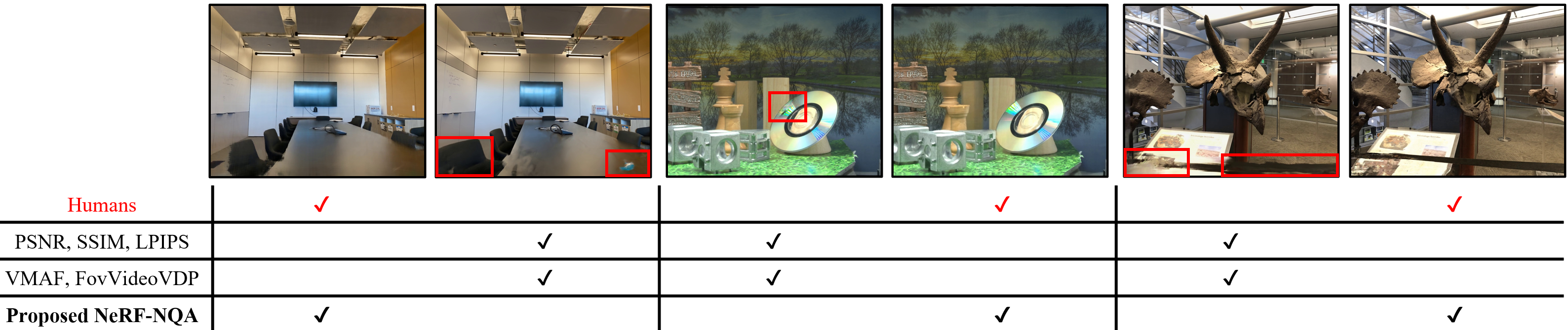}
  \caption{\textbf{Which NVS-generated scene (left or right) is better?} The areas manifesting significant blur and artifacts are demarcated with red boxes for enhanced visibility. In each instance, image quality assessment methods (PSNR, SSIM, LPIPS) and video quality assessment methods (VMAF, FovVideoVDP) diverge from human evaluations. Remarkably, the decisions from proposed quality assessment method exhibit strong concordance with human subjective perception.}
  \label{fig:teaser}
}

\graphicspath{{figs/}{figures/}{pictures/}{images/}{./}} 

\usepackage{tabu}                      
\usepackage{booktabs}                  
\usepackage{lipsum}                    
\usepackage{mwe}                       

\usepackage{mathptmx}                  

\begin{document}


\definecolor{ReviseColor}{RGB}{0,0,255}
\definecolor{DefaultColor}{RGB}{0,0,0}

\colorlet{HighlightColor}{DefaultColor}

\firstsection{Introduction}


\maketitle

The synthesis of photorealistic free views plays a pivotal role in enhancing user experiences in Virtual Reality (VR) and Augmented Reality (AR)~\cite{andersen2018ar, subramanyam2020comparing, whitlock2020graphical}. Such realistic rendering immerses users deeply into the VR or AR environment, making it easier for them to engage in the virtual content~\cite{gruber2014efficient, song2023nerfplayer}. In AR, the seamless integration of virtual objects with real-world scenes is vital, and photorealistic rendering ensures that these virtual elements appear natural and believable. In VR, efficient view synthesis techniques can generate these ``realistic'' views without extensive data storage for every perspective, optimizing application performance~\cite{poullis2008rapid, wang2022nerfcap}. The adaptability of these views to real-world lighting conditions ensures that virtual objects reflect, refract, and cast shadows realistically, enhancing the immersive experience.

However, the synthesis of photorealistic free views from limited RGB images collected from sparse viewpoints remains a pivotal challenge in the field of image-based rendering~\cite{li2007image, colburn2012image, hauswiesner2013virtual}. Recently, Neural View Synthesis (NVS) via implicit representations has emerged as a promising research field, with techniques such as Neural Radiance Fields (NeRF) \cite{mildenhall2020nerf} and its variants \cite{barron2022mip, sun2022direct, fridovich2022plenoxels, wizadwongsa2021nex, suhail2022light} gaining considerable attention for their exceptional fidelity and robustness. However, the quality assessment of NVS-generated scenes presents a complex task as it necessitates a comprehensive evaluation encompassing various dimensions, such as spatial fidelity and smoothness across consecutive views. This complexity is further amplified in immersive VR/AR environments, where users have the liberty to perceive NVS-generated scenes from unrestricted viewpoints~\cite{subramanyam2020comparing, whitlock2020graphical}.

Current quality assessment protocols for NVS-generated scenes are typically based on full-reference image quality assessment methods, such as PSNR, SSIM \cite{wang2004image}, and LPIPS \cite{zhang2018unreasonable}, on a subset of hold-out views. Nevertheless, these methods are primarily tailored for images, thus may not adequately capture the comprehensive and immersive quality of NVS-generated scenes as perceived by human observers. Figure~\ref{fig:teaser} illustrates such examples where the quality assessment methods diverge from human assessments. Along with the issue above, the absence of ground truth views from diverse viewpoints makes the comprehensive quality assessment even more challenging. For example, existing datasets like LLFF~\cite{mildenhall2019local} and DTU~\cite{jensen2014large} provide only images from sparse views, and even the datasets with reference videos~\cite{knapitsch2017tanks, chang2017matterport3d, jiang2022neuman} are often limited to fixed capturing paths, rendering them inadequate for assessing NVS methods with full-reference methods, as NVS is able to generate unlimited views.

\begin{figure}[htb]
  \centering
   \includegraphics[width=\linewidth]{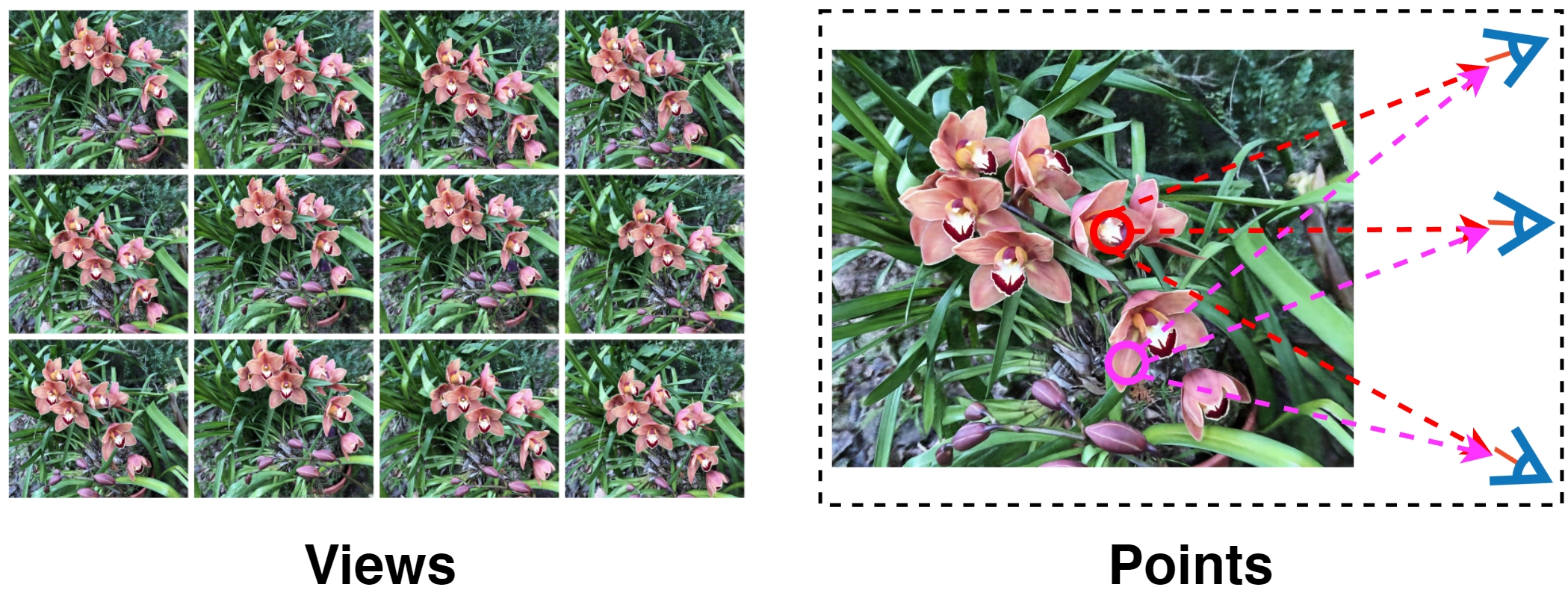}
   \caption{\textbf{NVS-generated scenes can be conceptualized from two perspectives: views (left) and points (right).} From the perspective of views, a scene can be perceived as an ensemble of views originating from diverse viewpoints. From the perspective of points, a scene can be perceived as a collection of surface points where each surface point can be observed from multiple angles.}
   \label{fig:views_and_points}
\end{figure}

To design a quality assessment method tailored for NVS-generated scenes, it is imperative to understand the foundation of these scenes. NVS-generated scenes can be conceptualized from two perspectives: views and points. Intuitively, an NVS-generated scene can be perceived as an ensemble of views originating from diverse viewpoints, as illustrated on the lefthand side of Figure~\ref{fig:views_and_points}. Predominant quality assessment methods, such as PSNR and SSIM~\cite{wang2004image}, are the view-centric approaches. These methods evaluate the quality of an NVS-generated scene by comparing, subsequently averaging the quality scores across all views to derive a final assessment. However, the view-centric methodologies have two inherent limitations. First, as previously highlighted, there is often a scarcity or complete absence of reference views in real-world scenarios. Second, a mere aggregation of quality scores might not accurately reflect the scene's quality as perceived by the human visual system. An alternative perspective treats an NVS-generated scene as a collection of surface points, as depicted on the righthand side of Figure~\ref{fig:views_and_points}. Given that each surface point can be observed from different view angles, the angular quality can be meticulously evaluated by scrutinizing the visual patterns associated with each surface point from varied orientations.

\textcolor{HighlightColor}{To bridge the gap, we introduce NeRF-NQA, the first no-reference quality assessment framework for synthesized scenes with dense viewpoints.} NeRF-NQA adopts a joint assessment approach consisting of both viewwise and pointwise assessment modules. The viewwise module evaluates the spatial quality of individual synthesized views, while the pointwise module focuses on the angular quality of individual scene surface point. Upon extensive evaluation and comparison against 23 established visual quality assessment methods, NeRF-NQA demonstrates superior performance on assessing the quality of NVS-synthesized scenes and better matches the perceptual judgement of human compared with existing assessment methods. The results shown in Figure~\ref{fig:teaser} illustrates the alignment of NeRF-NQA with human perceptual judgments, in contrast to mainstream image and video quality assessment methods. \textcolor{HighlightColor}{While our research primarily focuses on scenes synthesized by NeRF or other NVS variants, it is important to note that NeRF-NQA is designed with versatility in mind. It is applicable to any novel viewpoint synthesis method that provides dense viewpoints. Our focus on NVS is driven by its capability to reconstruct high-quality scenes, offering a wide array of viewpoints and rich diversity.} The primary contributions of this research are as follows:

\begin{itemize}

\item We propose \textcolor{HighlightColor}{the first no-reference quality assessment method for synthesized scenes with dense viewpoints,} considering the limited availability or absence of reference views in NVS-synthesized views.

\item We propose a joint quality assessment strategy, integrating both viewwise and pointwise approaches, to assess the quality of NVS-generated scenes. The viewwise approach focuses assessing the overall spatial quality of individual synthesized views and their inter-view consistency, while the pointwise approach focuses on the angular qualities, making it the pioneering approach on evaluating the quality of individual scene surface points and their compound inter-point quality.

\item To achieve accurate quality assessment without reference, we design a deep learning-based model for NeRF-NQA. Our extensive evaluation, comparing NeRF-NQA against 23 well-established visual quality assessment methods, clearly demonstrates its superiority over these traditional approaches by a substantial margin.

\end{itemize}

\section{Related Work}
\label{sec:related_work}

Quality assessment methods can be broadly classified into full-reference and no-reference, contingent upon the dependence of reference media~\cite{qu2021light}. Full-reference methods necessitate complete access to the reference media during quality score prediction. In contrast, no-reference methods determine quality without referencing the original media. The no-reference methods, while more intricate in design, are better suited for practical scenarios~\cite{qu2023lfacon}. This is particularly relevant for NVS quality assessment, given that some datasets, like LLFF, offer sparse images, rendering them inadequate for full-reference evaluations~\cite{mildenhall2019local}. Therefore, our research emphasizes no-reference quality assessment.

\noindent{\textbf{Image Quality Assessment.}}
The domain of image quality assessment is extensively studied. A plethora of full-reference methods for 2D images, such as PSNR, SSIM~\cite{wang2004image}, MS-SSIM~\cite{wang2003multiscale}, IW-SSIM~\cite{wang2010information}, VIF~\cite{sheikh2006image}, FSIM~\cite{zhang2011fsim}, GMSD~\cite{xue2013gradient}, VSI~\cite{zhang2014vsi}, DSS~\cite{balanov2015image}, HaarPSI~\cite{reisenhofer2018haar}, MDSI~\cite{nafchi2016mean}, LPIPS~\cite{zhang2018unreasonable}, PieAPP~\cite{prashnani2018pieapp}, and DISTS~\cite{ding2020image}, have been delineated in literature. PSNR is a prevalent objective quality assessment method that quantifies the quality of reconstructed images by comparing the maximum possible power of a signal to the power of corrupting noise, with a higher PSNR indicating a closer resemblance to the original image. In contrast, SSIM evaluates the perceptual quality of images by considering changes in structural information, luminance, and texture, providing a more comprehensive understanding of perceived image quality~\cite{wang2004image}. VIF gauges image quality by considering the mutual information shared between the reference and the distorted image, offering a nuanced assessment by accounting for characteristics of the human visual system~\cite{zhang2014vsi}. Lastly, the LPIPS employs deep learning techniques to measure perceptual differences between images, capturing intricate visual discrepancies that traditional methods might overlook~\cite{zhang2018unreasonable}. No-reference image quality assessment methods include the likes of BRISQUE~\cite{mittal2012no}, NIQE~\cite{mittal2012making}, and CLIP-IQA~\cite{wang2023exploring}. As one of the most popular, BRISQUE leverages the scene statistics of locally normalized luminance coefficients to measure potential reductions in "naturalness" due to distortions~\cite{mittal2012no}.

\noindent{\textbf{Video Quality Assessment.}} 
Beyond standard images, quality assessment methodologies exist for alternative visual media formats such as videos. Leading video quality methods encompass STRRED~\cite{soundararajan2012video}, VIIDEO~\cite{mittal2015completely}, VMAF~\cite{li2016toward}, and FovVideoVDP~\cite{mantiuk2021fovvideovdp}. STRRED focuses on the structural retention in videos, offering insights into the preservation of inherent video patterns post-processing~\cite{soundararajan2012video}. The VIIDEO, on the other hand, is a no-reference video quality assessment method that relies solely on the video being evaluated, utilizing intrinsic statistical regularities observed in natural videos~\cite{mittal2015completely}. VMAF, or Video Multi-Method Assessment Fusion, combines multiple algorithms to predict video quality, aligning closely with human perception by considering factors like texture, luminance, and motion~\cite{li2016toward}. Meanwhile, FovVideoVDP is a sophisticated method tailored for video quality assessment, taking into account the viewer's field of view to provide a more contextual evaluation~\cite{mantiuk2021fovvideovdp}. The video quality assessment techniques are adaptable to NVS, given that the synthesized view sequence can be analogously interpreted as a video.

\noindent{\textbf{Light-Field Quality Assessment.}}
Besides videos, light-field images are another media format that contains unique angular dimension for visual content. For light-field quality assessment, cutting-edge methods such as ALAS-DADS~\cite{qu2021light} and LFACon~\cite{qu2023lfacon} are the-state-of-the-arts in the field. ALAS-DADS is a pioneering no-reference light-field image quality assessment method designed for immersive media services. It introduces the light-field depthwise separable convolution for efficient spatial feature extraction and the light-field anglewise separable convolution to capture both spatial and angular features, ensuring a comprehensive yet efficient quality assessment~\cite{qu2021light}. LFACon, on the other hand, addresses light-field imaging's unique challenges by introducing the ``anglewise attention'' concept. This approach integrates a multihead self-attention mechanism into the angular domain of light-field images. With innovative attention kernels like anglewise self-attention, grid attention, and central attention, LFACon effectively gauges light-field image quality while optimizing computational efficiency~\cite{qu2023lfacon}. Those light-field quality assessment methodologies are compatible with NVS, as the synthesized views can be systematically rearranged into a light-field subview matrix, aligned with the respective camera poses of the views.

\noindent{\textbf{NVS Quality Assessment.}}
Presently, the evaluation of NVS methods or NeRF variants predominantly employs full-reference image quality methods~\cite{mildenhall2020nerf, barron2022mip, sun2022direct, fridovich2022plenoxels, wizadwongsa2021nex, suhail2022light}, which involve comparing the test set with the synthesized set image by image. In particular, PSNR and SSIM are the predominant image similarity methods, while LPIPS stands out as the leading perceptual deep-learned quality assessment method.
In this work, we assess the efficacy of the aforementioned quality assessment methods, including image, video, and light-field evaluation methods, to establish a performance benchmark for NVS quality assessment.

\begin{figure}[thb]
  \centering
   \includegraphics[width=\linewidth]{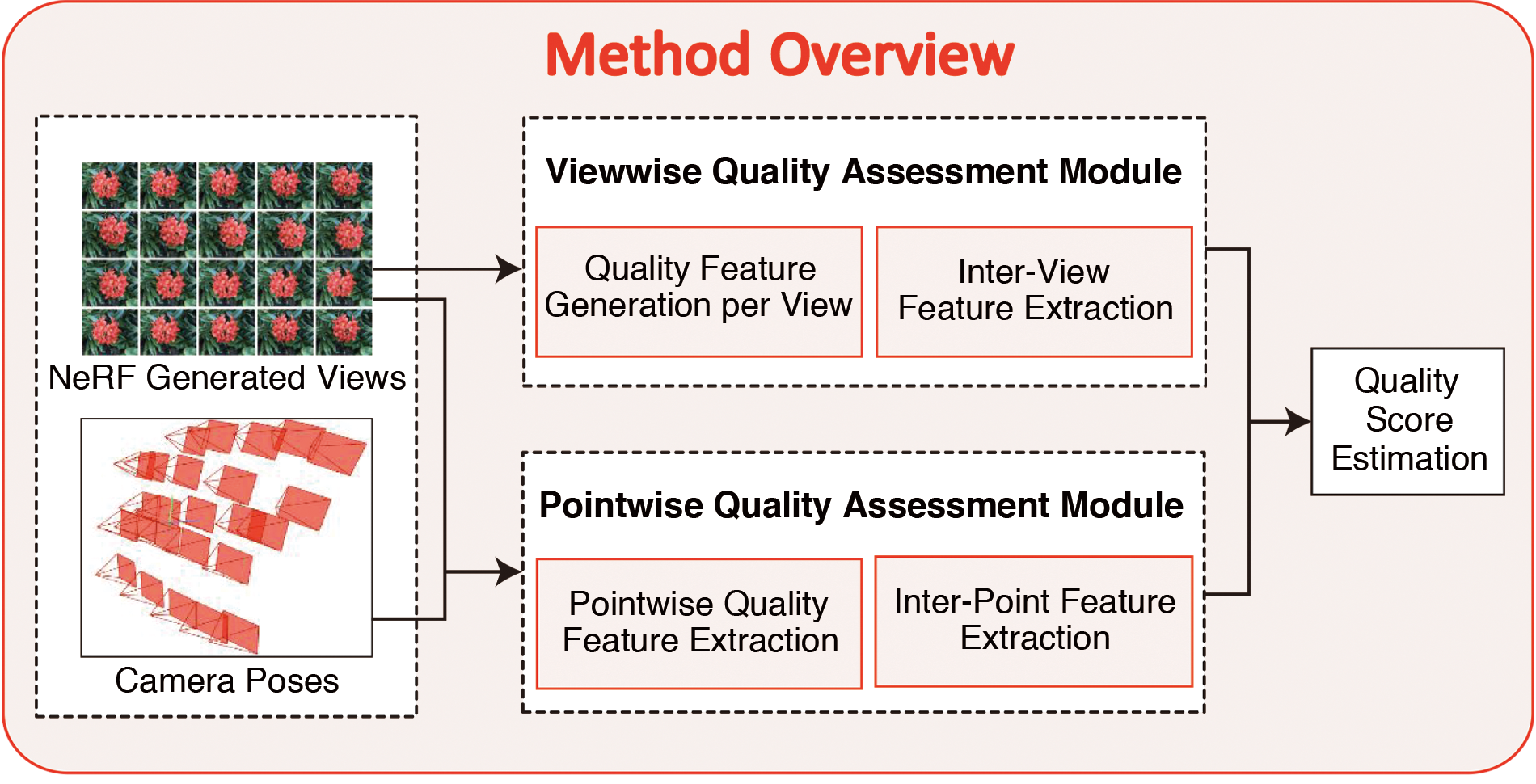}
    \vspace{-1.2em}
   \caption{{\bf Overview of the Proposed NVS Quality Assessment Framework.} }
   \vspace{-1.2em}
   \label{fig:framework}
\end{figure}

\section{Methodology}
\label{sec:method}

\subsection{Overview of NeRF-NQA}

As depicted in Figure~\ref{fig:framework}, the architecture of NeRF-NQA is principally divided into three major components: the Viewwise Quality Assessment Module, the Pointwise Quality Assessment Module, and the Quality Score Estimation Module.

The Viewwise Quality Assessment Module is designed to evaluate the spatial quality of scenes generated from NVS. This module ingests the synthesized views and undergoes two primary stages: Quality Feature Generation per View and Inter-View Feature Extraction. The output consists of viewwise quality features that encapsulate the spatial characteristics of the scene (detailed in Section~\ref{subsec:viewwise}).

The Pointwise Quality Assessment Module aims to capture angular quality features that are challenging for the Viewwise Module to assess.  Both NVS-generated views and their corresponding camera poses are taken as input and processed through a sequence of operations, including Pointwise Quality Feature Extraction and Inter-Point Feature Extraction, to yield pointwise quality features (detailed in Section~\ref{subsec:pointwise}).

Finally, the Quality Score Estimation Module employs a Multi-Layer Perceptron (MLP) to fuse the viewwise and pointwise features generated by the preceding modules, resulting in the final quality scores to offer a comprehensive assessment of the NVS scene. 
The intentional use of the MLP fusion aims to highlight the effectiveness of our proposed features. This fusion, common in representation learning as shown in references~\cite{chen2020simple, he2020momentum}, allows us to demonstrate the strength and discriminative power of extracted features without the interference of complex fusion techniques.

\begin{figure}[thb]
  \centering
   \includegraphics[width=\linewidth]{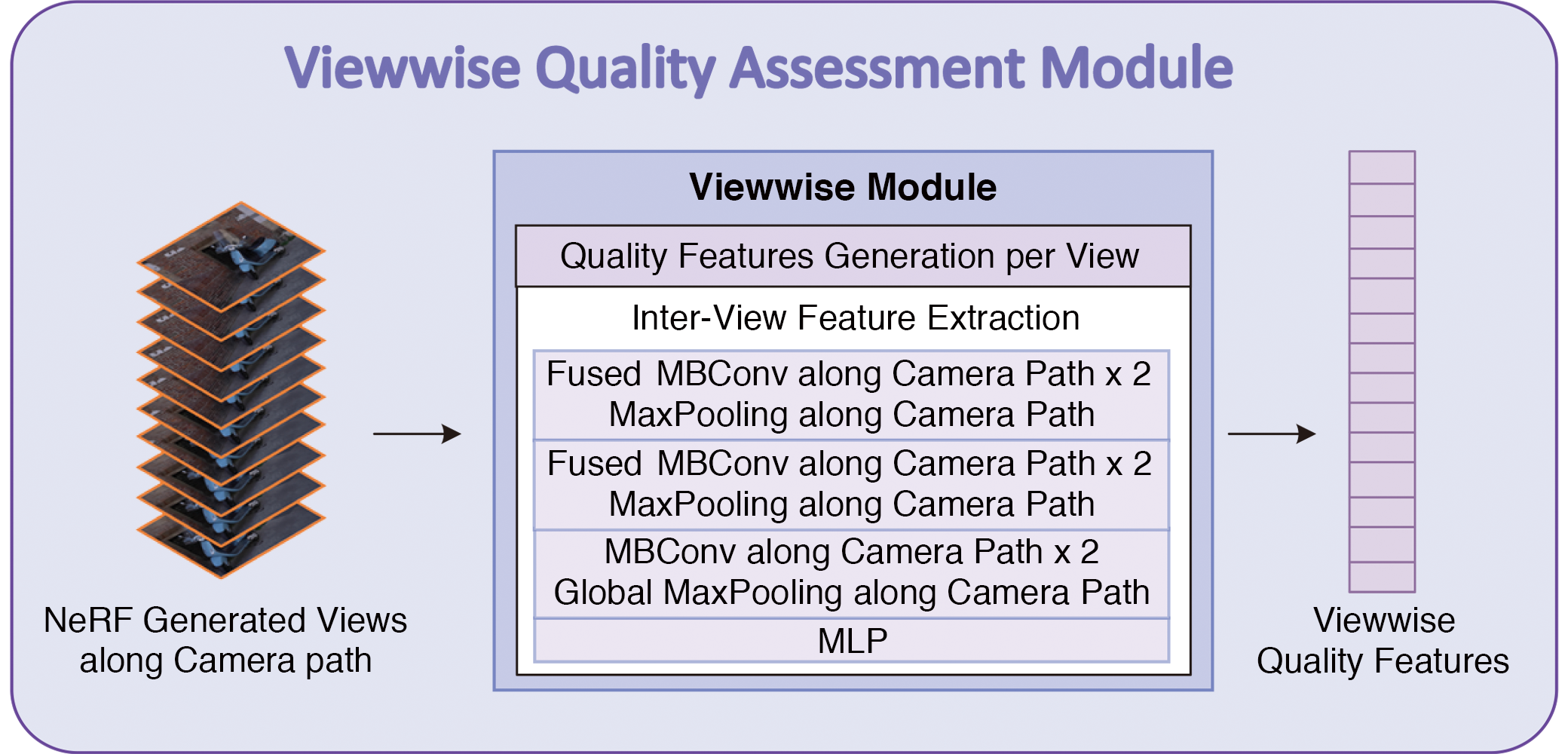}
    \vspace{-1.2em}
   \caption{{\bf The Structure of the Viewwise Quality Assessment Module.} }
   \vspace{-1.2em}
   \label{fig:viewwise}
\end{figure}

\subsection{Viewwise Quality Assessment}
\label{subsec:viewwise}

The quality of NVS-generated scene is intrinsically influenced by the quality of each synthesized view. After generating the quality features of individual views, it is imperative to holistically evaluate the final quality, factoring in the interrelation of these views. Given that NVS outcomes typically follow a camera trajectory, an intuitive approach is to analyze the quality features along this path.

Based on this concept, we introduce a viewwise quality assessment module, as depicted in Figure~\ref{fig:viewwise}. The module starts with an initial block, Quality Features Generation per View~\cite{mittal2011blind}, to individually assess the synthesized views to produce quality features for each view. Then, the Inter-View Feature Extraction block extracts features along the camera path, with the model structure inspired by EfficientNetV2~\cite{tan2021efficientnetv2}. Specifically, it integrates two repeated sets of two (Fused) MBConv layers (as per \cite{sandler2018mobilenetv2, tan2021efficientnetv2}) combined with MaxPooling, two standalone MBConv layers \cite{sandler2018mobilenetv2} followed by global MaxPooling and a MLP. The MBConv employs the inverted bottleneck structure \cite{sandler2018mobilenetv2} and depthwise convolutional layers \cite{howard2017mobilenets} to enhance memory efficiency. Additionally, a squeeze-and-excitation unit \cite{hu2018squeeze} is integrated within the MBConv to recalibrate channel-wise feature responses adaptively. The fused MBConv variant replaces depthwise convolutional layers with standard ones, proven to be more efficacious for larger spatial dimensions \cite{tan2021efficientnetv2}.All layers operate coherently along the camera path, allowing the viewwise module to integrate inter-view quality features. This ensures that the quality of each view is evaluated in conjunction with its neighboring views. The global MaxPooling layer ensures the module's compatibility with view sequences of varying lengths. 

\begin{figure*}[bht]
  \centering
   \includegraphics[width=1.0\linewidth]{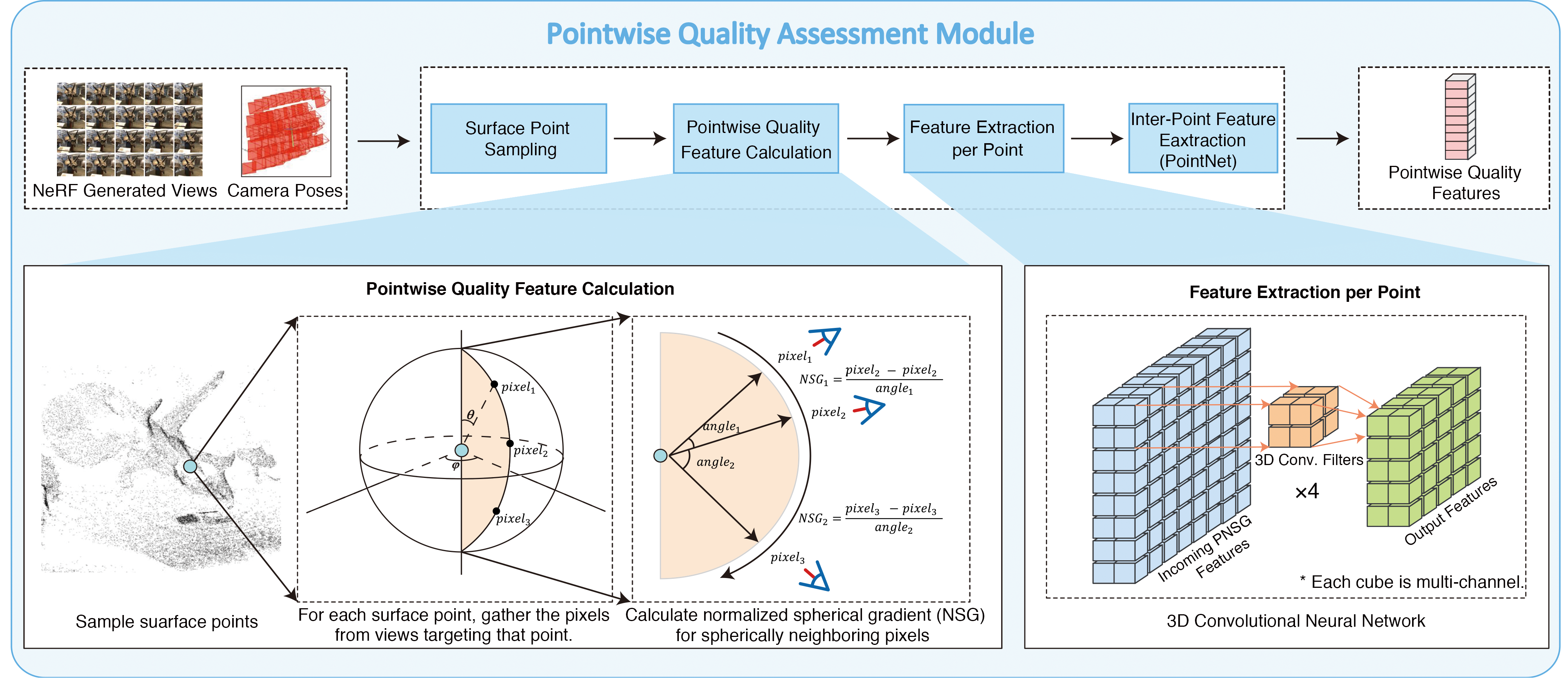}
    \vspace{-1.2em}
   \caption{{\bf The Detailed Architecture of the Pointwise Quality Assessment Module.} }
   \vspace{-1.2em}
   \label{fig:pointwise}
\end{figure*}

\subsection{Pointwise Quality Assessment}
\label{subsec:pointwise}

While the viewwise module adeptly captures the spatial quality of synthesized views, it encounters challenges in encapsulate the important angular quality explicitly. The angular quality, often delineated as the experience of observing a consistent location from varied angles~\cite{qu2023lfacon}, can be contextualized in NVS scenes as viewing a singular surface point from diverse viewpoints. To encapsulate the angular quality inherent in the NVS scenes, we introduce the pointwise quality assessment module.  This module is one of the key technical contribution of NeRF-NQA and its detailed design is shown in Figure~\ref{fig:pointwise}. The module commences by accepting NeRF synthesized views and camera poses, subsequently sampling sparse surface points via COLMAP~\cite{schonberger2016structure}. For each point, we compute pointwise quality features, elaborated in the subsequent paragraph. These high-dimensional pointwise quality features undergo further refinement in a feature extraction block, which distills the features per point and diminishes their dimensionalities. This block comprises four 3D convolutional layers followed by a MLP. Subsequently, an Inter-Point Feature Extraction block is designed by employing PointNet~\cite{qi2017pointnet} to extract inter-point quality features based on the spatial positioning of the points within the scene.

\noindent{\textbf{Pointwise Quality Feature Calculation.}}
To encapsulate the angular quality inherent to NVS scenes, we introduce the Pointwise Normalized Spherical Gradient map (PNSG) as the foundational pointwise quality features. The essence of PNSG lies in computing the gradient of pixel values observed from different viewpoints targeting at an identical surface point. The intricate procedures underpinning the pointwise quality feature calculation are delineated on the lower-left quadrant of Figure~\ref{fig:pointwise}. For each sampled surface point, we collate pixels from views targeting at the point. Subsequently, we compute the normalized spherical gradients (NSG) for spherically adjacent pixels. As depicted in the figure, for two pixels in proximity, the NSG is derived as the variance in pixel values normalized by the angular difference. Formally, let $o$ denotes a surface point, and $x_i, x_j$ are two pixels observing that point, NSG can be obtained by,

\begin{equation}
\label{eq:NSG}
\begin{aligned}
NSG(x_i, x_j) = \frac{I(x_i)-I(x_j)}{\sphericalangle x_i o x_j},
\end{aligned}
\end{equation}
where $I(x_i)$ and $I(x_j)$ denote the corresponding pixel values (i.e., vectors of RGB values), and $\sphericalangle x_i o x_j$ signifies the angular disparity between the two points.

\noindent{\textbf{Formal Definition of PNSG.}}
The PNSG is derived as an aggregation of NSG values. 
Consider a set of $n$ surface points, denoted as $\{ P_i\}^{n-1}_{i=0}$, for which we aim to compute the PNSG. For a given surface point $P_i$, we can collate pixels from all synthesized views targeting at that point, given the respective camera poses. Each pixel is associated with both its viewpoint position in 3D space and its RGB values. Subsequently, we transform the pixel positions from Cartesian to spherical coordinates, using the surface point as the spatial origin. This transformation allows us to represent the view direction of each pixel using azimuthal and polar angles.

Initially, we compute the NSG along the azimuthal axis by partitioning the polar axis into $b$ evenly spaced bins, represented as $\{ B_i\}^{b-1}_{i=0}$. Pixels are then grouped into the nearest bins. For a specific azimuthal bin $B_i$ containing $m_i$ pixels, we arrange the pixels by their azimuthal angles, denoted as $B_i = \{ x^i_j\}^{m_i-1}_{j=0}$. We then compute the NSG for each adjacent pair of pixels, resulting in $b$ bins of NSG along the azimuthal axis, represented as $NSG_{azi}$.

\begin{equation}
\label{eq:NSG_azi}
\begin{aligned}
NSG_{azi} = \{ \{ NSG(x^i_j, x^i_{j+1}) \}^{m_i-1}_{j=0} \}^{b-1}_{i=0}.
\end{aligned}
\end{equation}

In a similar vein, we compute the NSG along the polar axis, denoted as $NSG_{pol}$. The PNSG for the surface point $P_i$ is then represented as $\{NSG_{azi}^i, NSG_{pol}^i\}$. The cumulative PNSG for the entire scene is defined as:

\begin{equation}
\label{eq:PNSG}
\begin{aligned}
PNSG = \{ \{ NSG_{azi}^i, NSG_{pol}^i \}^{n-1}_{i=0} \}.
\end{aligned}
\end{equation}

From the derivations presented, it is apparent that the PNSG captures the dynamics within the angular domain, serving as a feature set for evaluating the angular quality inherent to NVS scenes.

\begin{table}[htb]
\caption{\textbf{Ablation study on effectiveness of NeRF-NQA variants (with or without pointwise module)} with quantitative evaluation (RMSE/SRCC) across the Fieldwork, LLFF, and Lab datasets. For each row, the best results are highlighted in bold.}
\label{tab:ablation}
\scriptsize
\renewcommand{\arraystretch}{1.0}
\setlength{\tabcolsep}{2pt}
\begin{tabular}{P{0.19\columnwidth} | P{0.12\columnwidth} P{0.11\columnwidth} | P{0.12\columnwidth} P{0.11\columnwidth} | P{0.12\columnwidth} P{0.11\columnwidth}}
\hline
\hline
NeRF-NQA &\multicolumn{2}{c}{Fieldwork}&\multicolumn{2}{c}{LLFF}&\multicolumn{2}{c}{Lab}\\
\cline{2-7}
Variant & RMSE  ↓ & SRCC ↑ & RMSE  ↓& SRCC ↑ & RMSE  ↓ & SRCC ↑ \\
\hline

w/o Pointwise & $\pmb{0.9202}$ & $0.9343$ & $0.8856$ & $0.7412$ & $1.0033$ & $0.8076$ \\
w/ Pointwise & $1.1969$ & $\pmb{0.9701}$ & $\pmb{0.5909}$ & $\pmb{0.9023}$ & $\pmb{0.6337}$ & $\pmb{0.8628}$ \\

\hline
\hline
\end{tabular}

\vspace{-1.2em}
\end{table}

\begin{table*}[htb]
\caption{\textbf{Quantitative evaluation of various quality assessment methods across the Fieldwork, LLFF, and Lab datasets}, using measures such as RMSE, SRCC, PLCC, and OR. For each column, the best results are highlighted in bold, with the last row indicating the enhancement relative to the second-best result. 
The results for full-reference video quality assessment methods are marked as "–" for the LLFF dataset due to the absence of ground-truth videos.}
\vspace{-0.6em}
\label{tab:benchmarking}
\scriptsize
\renewcommand{\arraystretch}{1.0}
\setlength{\tabcolsep}{2pt}
\begin{tabular}{P{0.06\textwidth}|P{0.125\textwidth}|P{0.058\textwidth}P{0.058\textwidth}P{0.058\textwidth}P{0.058\textwidth}|P{0.058\textwidth}P{0.058\textwidth}P{0.058\textwidth}P{0.058\textwidth}|P{0.058\textwidth}P{0.058\textwidth}P{0.058\textwidth}P{0.058\textwidth}}
\hline
\hline
&&\multicolumn{4}{c}{Fieldwork}&\multicolumn{4}{c}{LLFF}&\multicolumn{4}{c}{Lab}\\
\hline
Type & Method & RMSE  ↓ & SRCC ↑ & PLCC ↑& OR  ↓ & RMSE  ↓& SRCC ↑ & PLCC ↑ &OR  ↓ & RMSE  ↓ & SRCC ↑ & PLCC ↑ &OR  ↓ \\
\hline

\multirow{14}{*}{FR-IQA}&PSNR & $3.4726$ & $0.8941$ & $0.8609$ & $0.0000$ & $1.0871$ & $0.4058$ & $0.3931$ & $0.0000$ & $1.0633$ & $0.6090$ & $0.5250$ & $0.0000$\\
&SSIM & $2.4503$ & $0.9371$ & $0.9345$ & $0.0000$ & $1.0815$ & $0.4359$ & $0.4077$ & $0.0450$ & $1.0689$ & $0.5171$ & $0.3840$ & $0.1700$\\
&MS-SSIM & $2.8353$ & $0.9337$ & $0.9236$ & $0.0175$ & $1.0961$ & $0.3898$ & $0.3838$ & $0.0400$ & $1.1061$ & $0.4973$ & $0.2942$ & $0.1475$\\
&IW-SSIM & $2.4700$ & $0.9447$ & $0.9348$ & $0.0000$ & $1.0846$ & $0.4801$ & $0.4674$ & $0.0150$ & $1.2668$ & $0.5330$ & $0.2622$ & $0.1500$\\
&VIF & $3.3400$ & $0.9206$ & $0.9300$ & $0.0000$ & $1.1744$ & $0.1403$ & $0.1303$ & $0.0000$ & $1.1971$ & $0.5433$ & $0.3328$ & $0.0000$\\
&FSIM & $3.0174$ & $0.9332$ & $0.9304$ & $0.0000$ & $1.0872$ & $0.4502$ & $0.4250$ & $0.0050$ & $1.1181$ & $0.5239$ & $0.3182$ & $0.1275$\\
&GMSD & $3.4804$ & $0.9270$ & $0.9068$ & $0.0000$ & $1.0697$ & $0.4542$ & $0.4473$ & $0.0050$ & $1.0909$ & $0.5257$ & $0.3554$ & $0.0000$\\
&VSI & $3.7439$ & $0.7077$ & $0.7583$ & $0.0025$ & $1.1790$ & $0.1703$ & $0.1746$ & $0.0025$ & $1.3812$ & $0.2198$ & $0.1851$ & $0.0200$\\
&DSS & $2.4960$ & $0.9293$ & $0.8930$ & $0.0000$ & $0.9216$ & $0.6166$ & $0.6077$ & $0.0250$ & $1.1328$ & $0.5477$ & $0.4338$ & $0.0000$\\
&HaarPSI & $2.9704$ & $0.9412$ & $0.9298$ & $0.0000$ & $1.0612$ & $0.4772$ & $0.4651$ & $0.0000$ & $1.2917$ & $0.5485$ & $0.3418$ & $0.0000$\\
&MDSI & $3.1141$ & $0.9362$ & $0.9109$ & $0.0000$ & $1.0579$ & $0.4554$ & $0.4581$ & $0.0000$ & $1.1171$ & $0.5499$ & $0.4224$ & $0.0000$\\
&LPIPS & $2.6378$ & $0.8894$ & $0.9256$ & $0.0000$ & $1.1561$ & $0.1532$ & $0.2242$ & $0.0275$ & $1.1342$ & $0.4010$ & $0.3572$ & $0.0450$\\
&PieAPP & $2.8222$ & $0.9006$ & $0.8527$ & $0.0000$ & $1.1331$ & $0.3289$ & $0.2941$ & $0.0500$ & $0.9576$ & $0.7442$ & $0.6315$ & $0.0250$\\
&DISTS & $2.3891$ & $0.9215$ & $0.9383$ & $0.0300$ & $1.0841$ & $0.4215$ & $0.3628$ & $0.0150$ & $1.0604$ & $0.4400$ & $0.4164$ & $0.0175$\\
\hline
\multirow{3}{*}{NR-IQA}&BRISQUE & $3.8765$ & $-0.3824$ & $-0.4039$ & $0.0050$ & $1.2805$ & $0.1243$ & $0.1160$ & $0.0000$ & $1.1461$ & $0.3850$ & $0.3307$ & $0.0000$\\
&NIQE & $3.7584$ & $0.1934$ & $0.2077$ & $0.0700$ & $1.5258$ & $-0.0767$ & $-0.0740$ & $0.0000$ & $1.5213$ & $-0.3272$ & $-0.2698$ & $0.0000$\\
&CLIP-IQA & $2.9339$ & $0.7096$ & $0.7412$ & $0.0000$ & $1.6296$ & $-0.1653$ & $-0.1586$ & $0.0000$ & $1.6787$ & $-0.4358$ & $-0.3698$ & $0.0000$\\
\hline
\multirow{4}{*}{VQA}&STRRED & $1.7864$ & $0.9416$ & $0.8676$ & $0.1000$ & $-$ & $-$ & $-$ & $-$ & $1.0835$ & $0.5338$ & $0.4690$ & $0.0175$\\
&VMAF & $3.7054$ & $0.9142$ & $0.8987$ & $0.0000$ & $-$ & $-$ & $-$ & $-$ & $1.1395$ & $0.5077$ & $0.3200$ & $0.1050$\\
&FovVideoVDP & $3.8254$ & $0.4979$ & $0.4967$ & $0.0100$ & $-$ & $-$ & $-$ & $-$ & $1.3689$ & $0.1280$ & $0.1193$ & $0.0050$\\
&VIIDEO & $3.7445$ & $0.2902$ & $0.3640$ & $0.0000$ & $1.2384$ & $0.3213$ & $0.3077$ & $0.0000$ & $1.0526$ & $0.5677$ & $0.5134$ & $0.0000$\\
\hline
\multirow{2}{*}{LFIQA}&ALAS-DADS & $2.8179$ & $0.5223$ & $0.6299$ & $0.1050$ & $1.2504$ & $0.5498$ & $0.4691$ & $0.0225$ & $1.3447$ & $0.3367$ & $0.3936$ & $0.0000$\\
&LFACon & $3.1131$ & $0.4606$ & $0.5628$ & $0.0950$ & $0.9074$ & $0.5573$ & $0.6466$ & $0.3100$ & $0.7918$ & $0.5870$ & $0.7309$ & $0.0500$\\
\hline

&\textbf{NeRF-NQA}  & $\pmb{ 1.1969 }$ & $\pmb{ 0.9701 }$ & $\pmb{ 0.9804 }$ & $\pmb{ 0.0000 }$ & $\pmb{ 0.5909 }$ & $\pmb{ 0.9023 }$ & $\pmb{ 0.8858 }$ & $\pmb{ 0.0000 }$ & $\pmb{ 0.6337 }$ & $\pmb{ 0.8628 }$ & $\pmb{ 0.8720 }$ & $\pmb{ 0.0000 }$\\
&\textbf{Boost v.s. 2nd Best}  & $ \pmb{ +33.0\% } $ & $ \pmb{ +0.025 } $ & $ \pmb{ +0.042 } $ & $ \pmb{ - } $ & $ \pmb{ +34.9\% } $ & $ \pmb{ +0.286 } $ & $ \pmb{ +0.239 } $ & $ \pmb{ - } $ & $ \pmb{ +20.0\% } $ & $ \pmb{ +0.119 } $ & $ \pmb{ +0.141 } $ & $ \pmb{ - } $\\

\hline
\hline
\end{tabular}
\vspace{-1.5em}
\end{table*}

\section{Experiments}

\subsection{Datasets for Evaluation}

We evaluate and compare our proposed NeRF-NQA with existing quality assessment methods on three NVS datasets: Lab~\cite{liang2023perceptual}, LLFF~\cite{mildenhall2019local}, and Fieldwork~\cite{liang2023perceptual}. Lab dataset features 6 real scenes captured in a lab setting with a 2D gantry, facilitating both horizontal and vertical camera movements. Training views were taken on a uniform grid, and reference videos ranged from 300 to 500 frames~\cite{liang2023perceptual}. LLFF dataset comprises 8 real scenes captured via a handheld cellphone, each with sparse test views (20-30 images)~\cite{mildenhall2019local}. Poses for these images were computed using the COLMAP structure from motion~\cite{schonberger2016structure}. Fieldwork dataset contains 9 real scenes from outdoor urban areas and indoor museum spaces. These scenes are challenging due to intricate backgrounds, occlusions, and varying lighting. Reference videos typically have around 120 frames with diverse trajectories~\cite{liang2023perceptual}. For each dataset, we randomly designate four scenes for testing, while the remaining scenes are allocated for training. To mitigate overfitting, we conduct ten rounds of random surface sampling on every scene, effectively augmenting both the training and testing samples tenfold.

\begin{table*}[htbp]

\caption{\textbf{Comparative analysis of quality assessment methods for various evaluated scenes}, using RMSE and SRCC. The penultimate column presents the rankings of NeRF-NQA. The final column delineates either the enhancement achieved by NeRF-NQA over the second-leading method (when NeRF-NQA is top-ranked) or the difference relative to the foremost method (if NeRF-NQA doesn't achieve the best score). The results of VMAF are ``-'' for Flower, Fortress, Horns, and Room because these scenes have no ground-truth videos.}

\label{tab:scene_results}

\scriptsize
\renewcommand{\arraystretch}{1.2}
\setlength{\tabcolsep}{2pt}
\begin{tabular}{P{0.10\textwidth}|P{0.09\textwidth}|P{0.07\textwidth}P{0.06\textwidth}P{0.07\textwidth}P{0.07\textwidth}P{0.07\textwidth}P{0.09\textwidth}P{0.07\textwidth}P{0.08\textwidth}|P{0.04\textwidth}|P{0.09\textwidth}}
\hline
\hline
NVS Scene & Evaluation & PSNR & SSIM & LPIPS & BRISQUE & VMAF & VIIDEO & LFACon
 & \textbf{NeRF-NQA} & \textbf{Rank}& \textbf{Against Best Alt. Method}\\
\hline

\multirow{ 2 }{*}{Dinosaur} & RMSE ↓ & $2.9468$ & $2.5588$ & $2.5790$ & $3.4200$ & $3.2933$ & $3.2472$ & $2.9987$ & $0.5898$ & 1 & +77.0\% \\ \cline{2-2} \cline{11-12}
 & SRCC ↑ & $0.9522$ & $0.9344$ & $0.9338$ & $-0.2168$ & $0.9319$ & $0.6352$ & $0.4659$ & $0.9735$ & 1 & +0.021 \\ \cline{1-12}
\multirow{ 2 }{*}{Elephant} & RMSE ↓ & $1.5342$ & $1.6444$ & $1.5850$ & $2.0242$ & $1.6906$ & $1.8351$ & $1.3404$ & $0.9576$ & 1 & +28.6\% \\ \cline{2-2} \cline{11-12}
 & SRCC ↑ & $0.8930$ & $0.8402$ & $0.6311$ & $-0.2853$ & $0.8604$ & $0.4167$ & $0.6514$ & $0.9567$ & 1 & +0.064 \\ \cline{1-12}
\multirow{ 2 }{*}{Naiad-Sta.} & RMSE ↓ & $2.8302$ & $1.8489$ & $2.1174$ & $3.3172$ & $3.2272$ & $3.8593$ & $2.7868$ & $1.5221$ & 1 & +17.7\% \\ \cline{2-2} \cline{11-12}
 & SRCC ↑ & $0.9488$ & $0.9540$ & $0.8505$ & $-0.0667$ & $0.8184$ & $-0.3570$ & $0.4978$ & $0.9535$ & 2 & -0.001 \\ \cline{1-12}
\multirow{ 2 }{*}{Vespa} & RMSE ↓ & $5.4027$ & $3.3683$ & $3.7661$ & $5.7716$ & $5.5498$ & $5.2229$ & $4.4956$ & $1.4659$ & 1 & +56.5\% \\ \cline{2-2} \cline{11-12}
 & SRCC ↑ & $0.9449$ & $0.9621$ & $0.9530$ & $-0.0949$ & $0.9122$ & $-0.4887$ & $0.3489$ & $0.9708$ & 1 & +0.009 \\ \cline{1-12}
\multirow{ 2 }{*}{Flower} & RMSE ↓ & $1.2168$ & $1.2335$ & $1.4285$ & $1.2765$ & $-$ & $1.3719$ & $0.9715$ & $0.4967$ & 1 & +48.9\% \\ \cline{2-2} \cline{11-12}
 & SRCC ↑ & $0.4239$ & $0.3089$ & $-0.3086$ & $0.3279$ & $-$ & $0.1380$ & $0.5171$ & $0.9756$ & 1 & +0.458 \\ \cline{1-12}
\multirow{ 2 }{*}{Fortress} & RMSE ↓ & $0.9892$ & $0.8335$ & $0.8957$ & $1.3640$ & $-$ & $1.0133$ & $0.6981$ & $0.6458$ & 1 & +7.5\% \\ \cline{2-2} \cline{11-12}
 & SRCC ↑ & $0.3870$ & $0.7396$ & $0.6562$ & $-0.1220$ & $-$ & $0.2178$ & $0.6585$ & $0.8916$ & 1 & +0.152 \\ \cline{1-12}
\multirow{ 2 }{*}{Horns} & RMSE ↓ & $0.9626$ & $0.9777$ & $1.0308$ & $1.2459$ & $-$ & $1.1435$ & $0.9492$ & $0.4264$ & 1 & +55.1\% \\ \cline{2-2} \cline{11-12}
 & SRCC ↑ & $0.7518$ & $0.6839$ & $0.2275$ & $0.1698$ & $-$ & $0.4256$ & $0.5398$ & $0.8859$ & 1 & +0.134 \\ \cline{1-12}
\multirow{ 2 }{*}{Room} & RMSE ↓ & $1.1583$ & $1.2274$ & $1.2003$ & $1.2316$ & $-$ & $1.3848$ & $0.9807$ & $0.7422$ & 1 & +24.3\% \\ \cline{2-2} \cline{11-12}
 & SRCC ↑ & $0.3704$ & $0.4322$ & $0.2589$ & $-0.3193$ & $-$ & $-0.0756$ & $0.4285$ & $0.7995$ & 1 & +0.367 \\ \cline{1-12}
\multirow{ 2 }{*}{CD-Occ.} & RMSE ↓ & $0.9937$ & $1.1628$ & $1.1442$ & $1.0021$ & $0.9256$ & $0.8736$ & $0.5789$ & $0.4982$ & 1 & +13.9\% \\ \cline{2-2} \cline{11-12}
 & SRCC ↑ & $0.1293$ & $0.0756$ & $0.1718$ & $-0.4688$ & $0.0196$ & $-0.1964$ & $0.7807$ & $0.8439$ & 1 & +0.063 \\ \cline{1-12}
\multirow{ 2 }{*}{Animals} & RMSE ↓ & $1.3983$ & $1.3806$ & $1.3976$ & $1.2730$ & $1.2836$ & $1.2849$ & $0.7894$ & $0.9927$ & 2 & -20.5\% \\ \cline{2-2} \cline{11-12}
 & SRCC ↑ & $-0.0865$ & $0.0263$ & $0.0480$ & $-0.3974$ & $0.0052$ & $-0.4333$ & $0.7883$ & $0.7890$ & 1 & +0.001 \\ \cline{1-12}
\multirow{ 2 }{*}{Metal} & RMSE ↓ & $0.7097$ & $0.4987$ & $0.4562$ & $0.4904$ & $0.8212$ & $0.7927$ & $0.6188$ & $0.3854$ & 1 & +15.5\% \\ \cline{2-2} \cline{11-12}
 & SRCC ↑ & $0.5773$ & $0.0930$ & $0.0206$ & $0.1244$ & $0.1881$ & $0.2620$ & $0.4390$ & $0.4751$ & 2 & -0.102 \\ \cline{1-12}
\multirow{ 2 }{*}{Toys} & RMSE ↓ & $1.0374$ & $1.0312$ & $1.2941$ & $1.5456$ & $1.4194$ & $1.1788$ & $1.0802$ & $0.4733$ & 1 & +54.1\% \\ \cline{2-2} \cline{11-12}
 & SRCC ↑ & $0.6605$ & $0.2962$ & $0.3089$ & $-0.1676$ & $0.4893$ & $0.3961$ & $0.4146$ & $0.7574$ & 1 & +0.097 \\ \cline{1-12}

\hline
\hline
\end{tabular}
\vspace{-2.0em}
\end{table*}

\subsection{Perceptual Quality Labels}

The perceptual quality labels are derived from subjective experiments by Liang et al.~\cite{liang2023perceptual}. They engaged 39 color-normal volunteers, with each participant completing 4–5 batches of comparisons using ASAP~\cite{mikhailiuk2021active}. The results, scaled from pairwise comparisons, were articulated in Just-Objectionable-Difference (JOD) units via the Thurstone Case V observer model~\cite{perez2017practical}. The JOD scores are offset by reference scores and thus predominantly negative values. A JOD score of 0 indicates undistorted quality. Higher JOD values suggest better quality perceived by human visual systems.

These experiments encompassed ten representative NVS methods, showcasing a variety of models with both explicit and implicit geometric representations, different rendering models, and optimization strategies. NeRF~\cite{mildenhall2020nerf} introduces a neural volumetric representation optimized for image-based scene reconstruction and the synthesis of novel views.  Mip-NeRF~\cite{barron2022mip} offers a multiscale representation tailored for anti-aliasing in view synthesis. Both DVGO~\cite{sun2022direct} and Plenoxels~\cite{fridovich2022plenoxels} employ hybrid representations, streamlining the training and rendering processes. NeX~\cite{wizadwongsa2021nex} leverages multi-plane images combined with trainable basis functions, specifically designed to render view-dependent effects in forward-facing scenes. LFNR~\cite{suhail2022light} adopts a light-field representation, incorporating an epipolar constraint to enhance the rendering process. Furthermore, both IBRNet~\cite{wang2021ibrnet} and GNT~\cite{wang2022attention} are built upon the NeRF model and promote the generalizability. For IBRNet and GNT, both cross-scene models (GNT-C and IBRNet-C) and scene-specific models (GNT-S and IBRNet-S) were tested.

\begin{figure*}[bht]
\centering
\includegraphics[width=0.85\textwidth]{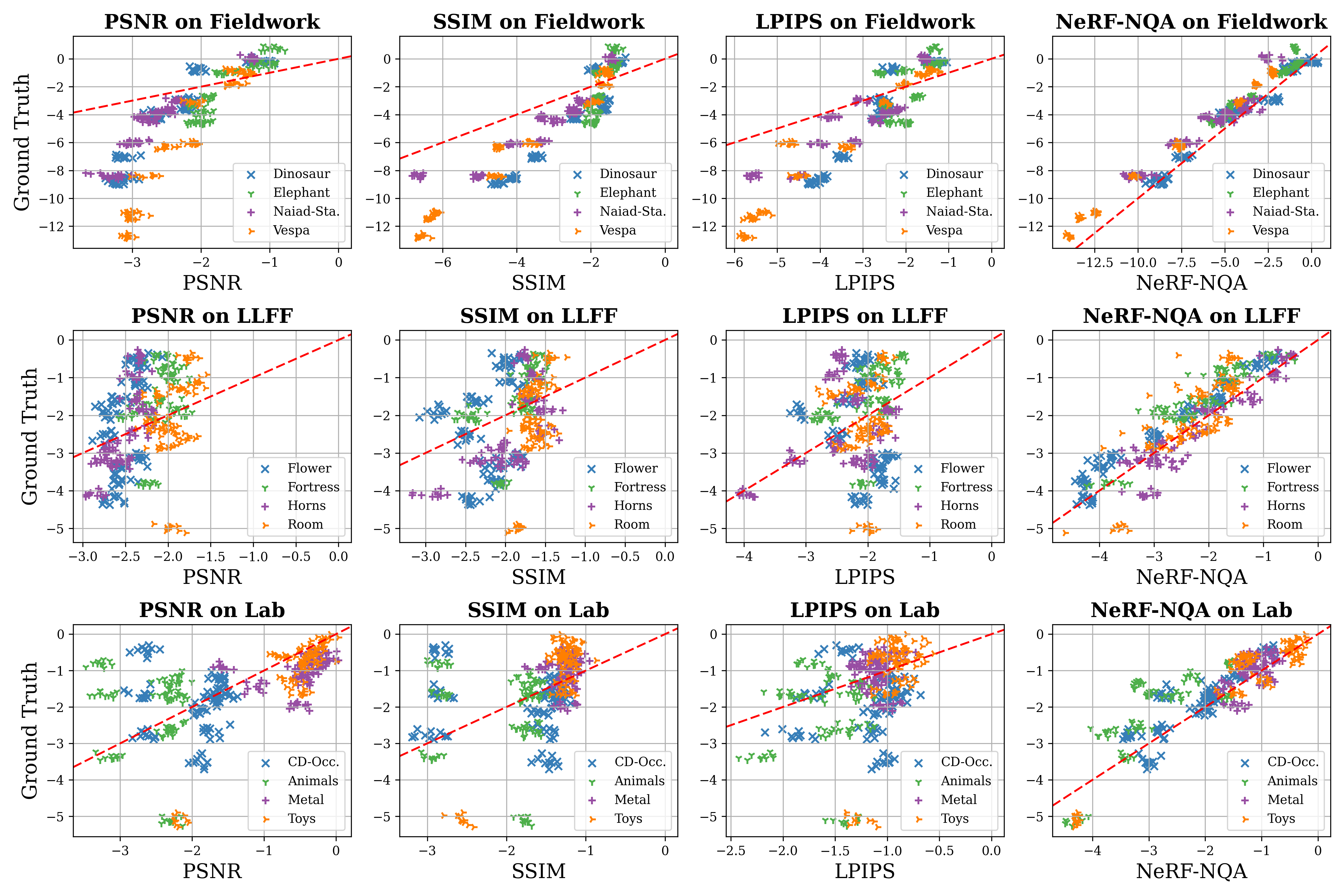}
\vspace{-0.6em}
\caption{\textbf{Scatter plots illustrating the correlation between ground truth JOD and estimation made by the most widely used metrics for NVS (i.e., PSNR, SSIM, and LPIPS) and the proposed NeRF-NQA across the Fieldwork, LLFF, and Lab datasets.} Distinct symbols and colors denote various scene. Each subfigure features a red line representing the ideal prediction trajectory (i.e., ground truth == metric estimation). Notably, proximity of data points to this red line signifies superior metric performance.}
\label{fig:bench_plots}
\vspace{-0.5em}
\end{figure*}

\subsection{Training Setup}

\textcolor{HighlightColor}{The model was trained utilizing the ADAM optimizer~\cite{kingma2014adam}, over 200 epochs with a batch size of 10. It is designed as a generalized model, which, post-training, is capable of operating across diverse scenes without necessitating scene-specific fine-tuning. The weights, established during this initial training phase, are maintained consistently. In other words, once the model is trained, it is supposed to be proficiently applied to unseen scenes across different datasets. The computational experiments were conducted on a desktop equipped with an AMD 5950X processor, an RTX 3090 GPU, and 32GB of RAM, operating on Windows 10. The implementation replied on the PyTorch~\cite{paszke2019pytorch}.}

\subsection{Metrics to Evaluate the Quality Assessment Methods}

In the realm of quality assessment, several metrics are commonly employed to quantify the performance of quality assessment methods~\cite{qu2021light, qu2023lfacon}. Among these, the Root Mean Square Error (RMSE)~\cite{dekking2005modern} serves as a standard measure of the differences between predicted and ground-truth values, with lower RMSE values indicating more accurate predictions. The Spearman Rank Order Correlation Coefficient (SRCC)~\cite{zwillinger1999crc} assesses the strength and direction of the monotonic relationship between the predicted and ground-truth scores. Higher SRCC values signify a stronger correlation and, consequently, better performance. Similarly, the Pearson Linear Correlation Coefficient (PLCC)~\cite{dekking2005modern} evaluates the linear correlation between the predicted and actual quality scores. A PLCC value closer to 1 indicates a strong positive linear correlation, thereby suggesting that the quality assessment algorithm is highly accurate in its predictions. Additionally, the Outlier Ratio (OR) is another important metric that is often calculated using statistical methods such as Tukey's fences~\cite{tukey1977exploratory}. OR measures the proportion of data points that deviate significantly from the rest of the data distribution, providing insights into the robustness of the algorithm against anomalies or extreme values. Lower OR values are indicative of fewer outliers and thus suggest a more reliable and consistent performance. Collectively, these metrics provide a comprehensive evaluation on the quality assessment method's performance in terms of both accuracy and correlation with human perceptual judgments.

\begin{table*}[htbp]

\caption{\textbf{Comparative analysis of quality assessment methods for different NVS methods}, using RMSE and SRCC. The last two columns show NeRF-NQA's ranking and its performance relative to the top or second-best method.}
\label{tab:nvs_method_results}


\scriptsize
\renewcommand{\arraystretch}{1.2}
\setlength{\tabcolsep}{2pt}
\begin{tabular}{P{0.10\textwidth}|P{0.09\textwidth}|P{0.07\textwidth}P{0.06\textwidth}P{0.07\textwidth}P{0.07\textwidth}P{0.07\textwidth}P{0.09\textwidth}P{0.07\textwidth}P{0.08\textwidth}|P{0.04\textwidth}|P{0.09\textwidth}}
\hline
\hline
NVS Method & Evaluation & PSNR & SSIM & LPIPS & BRISQUE & VMAF & VIIDEO & LFACon
 & \textbf{NeRF-NQA} & \textbf{Rank}& \textbf{Against Best Alt. Method}\\
\hline

\multirow{ 2 }{*}{DVGO} & RMSE ↓ & $1.1971$ & $1.3403$ & $1.4128$ & $1.4384$ & $1.5433$ & $1.3659$ & $0.9998$ & $0.6837$ & 1 & +31.6\% \\ \cline{2-2} \cline{11-12}
 & SRCC ↑ & $0.4763$ & $0.4317$ & $0.2963$ & $-0.1425$ & $0.1293$ & $0.4086$ & $0.4711$ & $0.8537$ & 1 & +0.377 \\ \cline{1-12}
\multirow{ 2 }{*}{GNT-C} & RMSE ↓ & $3.7748$ & $2.6914$ & $3.0733$ & $4.1168$ & $3.9801$ & $3.9746$ & $3.4138$ & $0.6789$ & 1 & +74.8\% \\ \cline{2-2} \cline{11-12}
 & SRCC ↑ & $0.6023$ & $0.8206$ & $0.6963$ & $0.1707$ & $-0.1342$ & $0.3088$ & $0.4432$ & $0.9604$ & 1 & +0.140 \\ \cline{1-12}
\multirow{ 2 }{*}{GNT-S} & RMSE ↓ & $3.1913$ & $2.2837$ & $2.3795$ & $3.4402$ & $3.3729$ & $3.2597$ & $2.7248$ & $1.0084$ & 1 & +55.8\% \\ \cline{2-2} \cline{11-12}
 & SRCC ↑ & $0.5641$ & $0.7092$ & $0.8658$ & $0.3071$ & $0.1773$ & $0.4936$ & $0.6731$ & $0.9229$ & 1 & +0.057 \\ \cline{1-12}
\multirow{ 2 }{*}{IBRNet-C} & RMSE ↓ & $2.8699$ & $2.1395$ & $2.3615$ & $3.0050$ & $3.0304$ & $3.0659$ & $2.6500$ & $0.7860$ & 1 & +63.3\% \\ \cline{2-2} \cline{11-12}
 & SRCC ↑ & $0.8464$ & $0.8922$ & $0.8832$ & $0.5059$ & $0.2711$ & $0.3542$ & $0.3833$ & $0.9398$ & 1 & +0.048 \\ \cline{1-12}
\multirow{ 2 }{*}{IBRNet-S} & RMSE ↓ & $1.8191$ & $1.3968$ & $1.4331$ & $2.0398$ & $2.1254$ & $2.0002$ & $1.2924$ & $0.8834$ & 1 & +31.6\% \\ \cline{2-2} \cline{11-12}
 & SRCC ↑ & $0.8623$ & $0.8841$ & $0.7367$ & $0.5380$ & $0.2084$ & $0.2214$ & $0.7325$ & $0.9189$ & 1 & +0.035 \\ \cline{1-12}
\multirow{ 2 }{*}{LFNR} & RMSE ↓ & $1.6693$ & $1.3496$ & $1.2549$ & $1.5739$ & $1.7614$ & $1.6298$ & $1.2313$ & $0.7623$ & 1 & +38.1\% \\ \cline{2-2} \cline{11-12}
 & SRCC ↑ & $0.2339$ & $0.3743$ & $0.8081$ & $0.6265$ & $0.1286$ & $0.3687$ & $0.7098$ & $0.9668$ & 1 & +0.159 \\ \cline{1-12}
\multirow{ 2 }{*}{MipNeRF} & RMSE ↓ & $1.0777$ & $1.0727$ & $1.1959$ & $2.0036$ & $1.6776$ & $1.7545$ & $1.0997$ & $1.0721$ & 1 & +0.1\% \\ \cline{2-2} \cline{11-12}
 & SRCC ↑ & $0.5050$ & $0.3566$ & $0.1638$ & $-0.2746$ & $0.4199$ & $0.4742$ & $0.0622$ & $0.5664$ & 1 & +0.061 \\ \cline{1-12}
\multirow{ 2 }{*}{NeRF} & RMSE ↓ & $1.9513$ & $1.1090$ & $1.2401$ & $2.2956$ & $2.2111$ & $2.1238$ & $1.9637$ & $0.9019$ & 1 & +18.7\% \\ \cline{2-2} \cline{11-12}
 & SRCC ↑ & $0.7639$ & $0.8048$ & $0.7820$ & $0.2147$ & $0.6347$ & $0.5441$ & $0.5653$ & $0.9326$ & 1 & +0.128 \\ \cline{1-12}
\multirow{ 2 }{*}{NeX} & RMSE ↓ & $1.2453$ & $1.2616$ & $1.2307$ & $1.5399$ & $1.1071$ & $1.4861$ & $1.0716$ & $0.8470$ & 1 & +21.0\% \\ \cline{2-2} \cline{11-12}
 & SRCC ↑ & $0.5811$ & $0.6754$ & $0.5128$ & $-0.0095$ & $0.7404$ & $0.3016$ & $0.4478$ & $0.8184$ & 1 & +0.078 \\ \cline{1-12}
\multirow{ 2 }{*}{Plenoxel} & RMSE ↓ & $1.0900$ & $1.0692$ & $1.0700$ & $1.3264$ & $1.4421$ & $1.1829$ & $0.7990$ & $0.8204$ & 2 & -2.6\% \\ \cline{2-2} \cline{11-12}
 & SRCC ↑ & $0.3497$ & $0.3211$ & $0.4910$ & $-0.2480$ & $-0.0534$ & $0.3419$ & $0.6916$ & $0.8570$ & 1 & +0.165 \\ \cline{1-12}

\hline
\hline
\end{tabular}
\vspace{-1.0em}
\end{table*}

\subsection{Ablation Study on the Design of NeRF-NQA Model}

As elaborated in Section~\ref{sec:method}, the Pointwise Module is specifically designed to capture angular quality features that are inherently difficult for the Viewwise Module for assessment. To empirically validate the efficacy of the Pointwise Module, we construct two variants of NeRF-NQA: one incorporating the Pointwise Module and the other excluding it. Comparative performance metrics for these variants are presented in Table~\ref{tab:ablation}. Our experimental findings reveal that the NeRF-NQA variant with the Pointwise Module consistently outperforms its counterpart across nearly all evaluation criteria, with the exception of RMSE on the Fieldwork dataset, where the results are closely aligned. Notably, in the LLFF dataset, the fully-equipped NeRF-NQA demonstrates a 33.3\% reduction on RMSE and a 0.1611 increase on SRCC. These outcomes substantiate the utility and effectiveness of the Pointwise Module, thereby justifying its inclusion in subsequent experiments.

\begin{figure}[htb]

\scriptsize
\renewcommand{\arraystretch}{1.2}
\setlength{\tabcolsep}{2pt}
\begin{tabular}{P{0.20\columnwidth}|P{0.18\columnwidth}|P{0.18\columnwidth}|P{0.18\columnwidth}|P{0.18\columnwidth}} 
\hline\hline

\multicolumn{5}{c}{\textbf{Scene [Dinosaur] Synthesized by NVS method (GNT-C)} }\\
\multicolumn{5}{c} {\includegraphics[width=0.98\columnwidth, height=0.28\columnwidth]{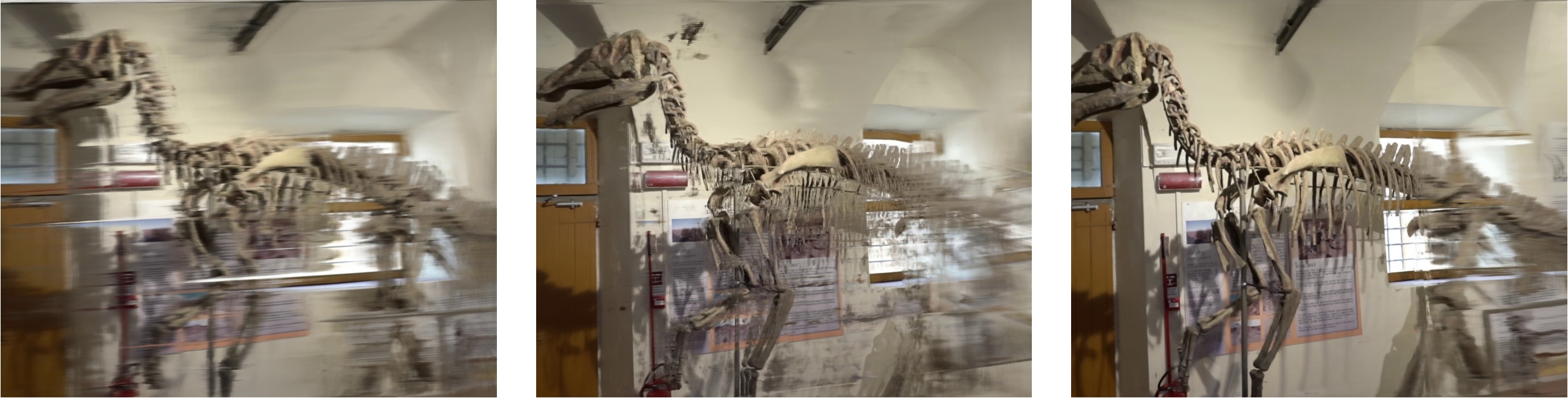}} \\
\multicolumn{5}{c}{Ground Truth JOD: \underline{-8.7655} }\\
\hline
Metric & PSNR & SSIM & LPIPS & \textbf{NeRF-NQA}\\ 
\hline
Estimation & -3.3529 & -4.4325 &  -4.2286 & $\pmb{-8.6999}$\\
\hline\hline
\hline\hline

\multicolumn{5}{c}{\textbf{Scene [Naiad-Sta.] Synthesized by NVS method (Nex)} }\\
\multicolumn{5}{c} {\adjustbox{valign=c,margin=0 1pt 0 1pt}{\includegraphics[width=0.98\columnwidth, height=0.28\columnwidth]{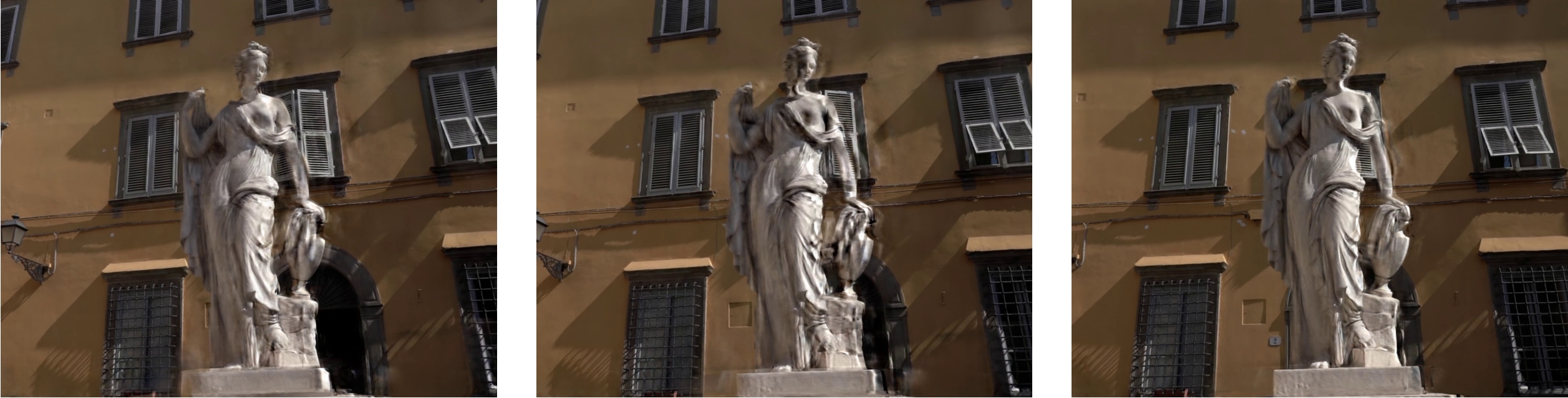}} }\\
\multicolumn{5}{c}{Ground Truth JOD: \underline{-3.6308} }\\
\hline
Metric & PSNR & SSIM & LPIPS & \textbf{NeRF-NQA}\\ 
\hline
Estimation & -2.5225 & -2.4639 &  -2.2143 & $\pmb{-3.5821}$\\
\hline\hline
\hline\hline

\multicolumn{5}{c}{\textbf{Scene [Horns] Synthesized by NVS method (MipNeRF)} }\\
\multicolumn{5}{c} {\adjustbox{valign=c,margin=0 1pt 0 1pt}{\includegraphics[width=0.98\columnwidth, height=0.28\columnwidth]{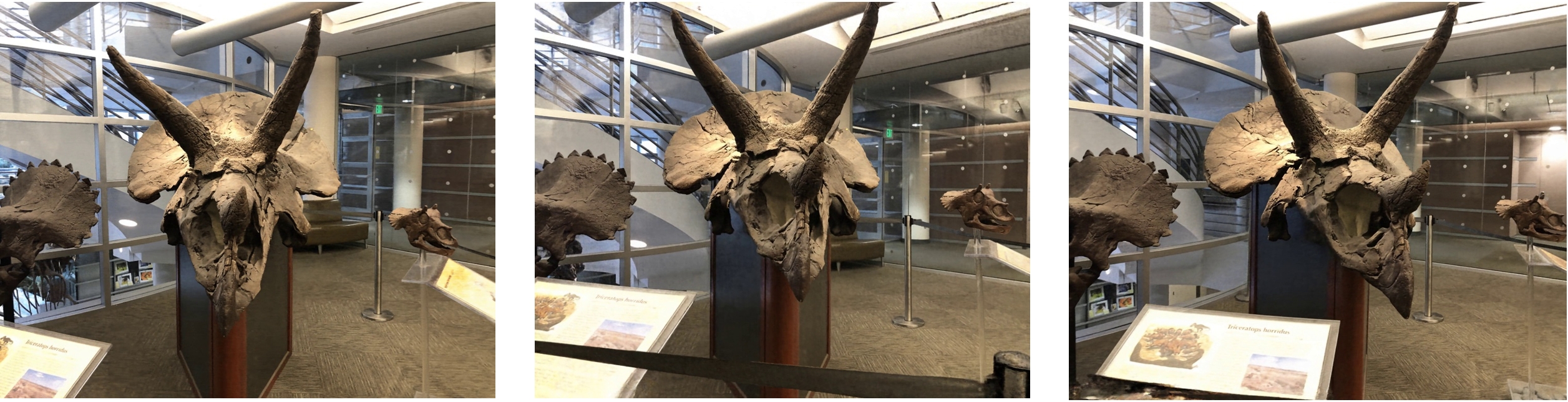}} }\\
\multicolumn{5}{c}{Ground Truth JOD: \underline{-0.4222} }\\
\hline
Metric & PSNR & SSIM & LPIPS & \textbf{NeRF-NQA}\\ 
\hline
Estimation & -2.4713 & -1.7241 &  -2.3615 & $\pmb{-0.4164}$\\
\hline\hline

\end{tabular}
\vspace{-0.6em}
\caption{\textbf{Illustration of sample scenes generated via various NVS methods, accompanied by the corresponding ground truth JOD}, which is underlined for emphasis. The estimations derived from NeRF-NQA are highlighted in bold and compared against those obtained from prevalent metrics for NVS, namely PSNR, SSIM, and LPIPS.}
\label{fig:example_scenes}
\vspace{-1.2em}
\end{figure}

\subsection{Comparison with Other Quality Assessment Methods}
Our benchmarking considered prevalent full-reference image quality assessment metrics (FR-IQA) such as PSNR, SSIM~\cite{wang2004image}, MS-SSIM~\cite{wang2003multiscale}, IW-SSIM~\cite{wang2010information}, VIF~\cite{sheikh2006image}, FSIM~\cite{zhang2011fsim}, GMSD~\cite{xue2013gradient}, VSI~\cite{zhang2014vsi}, DSS~\cite{balanov2015image}, HaarPSI~\cite{reisenhofer2018haar}, MDSI~\cite{nafchi2016mean}, LPIPS~\cite{zhang2018unreasonable}, PieAPP~\cite{prashnani2018pieapp}, and DISTS~\cite{ding2020image}, along with no-reference image quality assessment metrics (NR-IQA) such as BRISQUE~\cite{mittal2012no}, NIQE~\cite{mittal2012making}, and CLIP-IQA~\cite{wang2023exploring}. We also included video quality assessment methods (VQA) such as STRRED~\cite{soundararajan2012video}, VIIDEO~\cite{mittal2015completely}, VMAF~\cite{li2016toward}, and FovVideoVDP~\cite{mantiuk2021fovvideovdp}, and two state-of-the-art light-field quality assessment methods (LFIQA) including ALAS-DADS~\cite{qu2021light}, and LFACon~\cite{qu2023lfacon}. A detailed discussion of these quality metrics can be found in Section \ref{sec:related_work}.

As delineated in Table~\ref{tab:benchmarking}, the evaluation results (RMSE, SRCC, PLCC and OR) of the 24 quality assessment methods (including NeRF-NQA) are presented across the Fieldwork, LLFF, and Lab datasets. The best results for each column are accentuated in boldface, while the bottom row quantifies the relative improvement over the second-best method. Upon examination of the table, it is evident that the proposed NeRF-NQA method consistently outshines all other benchmarked methods. Specifically, on the Fieldwork dataset, NeRF-NQA exhibits a remarkable 33\% improvement on RMSE compared to the second-best method. In the LLFF dataset, NeRF-NQA demonstrates a 34.9\% enhancement on RMSE, a 0.286 increment on SRCC, and a 0.239 rise on PLCC over the second-best results. On the Lab dataset, NeRF-NQA achieves significant gains: a 20.0\% improvement in RMSE, a 0.119 increase in SRCC, and a 0.141 uptick in PLCC.

In summary, NeRF-NQA remarkably surpasses all other quality assessment methods, encompassing well-established FR-IQA methods such as PSNR, SSIM, and LPIPS, NR-IQA methods like BRISQUE and NIQE, recent advancements like CLIP-IQA, as well as VQA methods including VMAF and FovVideoVDP, and the-state-of-the-art LFIQA methods such as ALAS-DADS and LFACon.


\begin{figure}[bht]
\centering
\subfloat[]{
    \includegraphics[width=0.85\columnwidth]{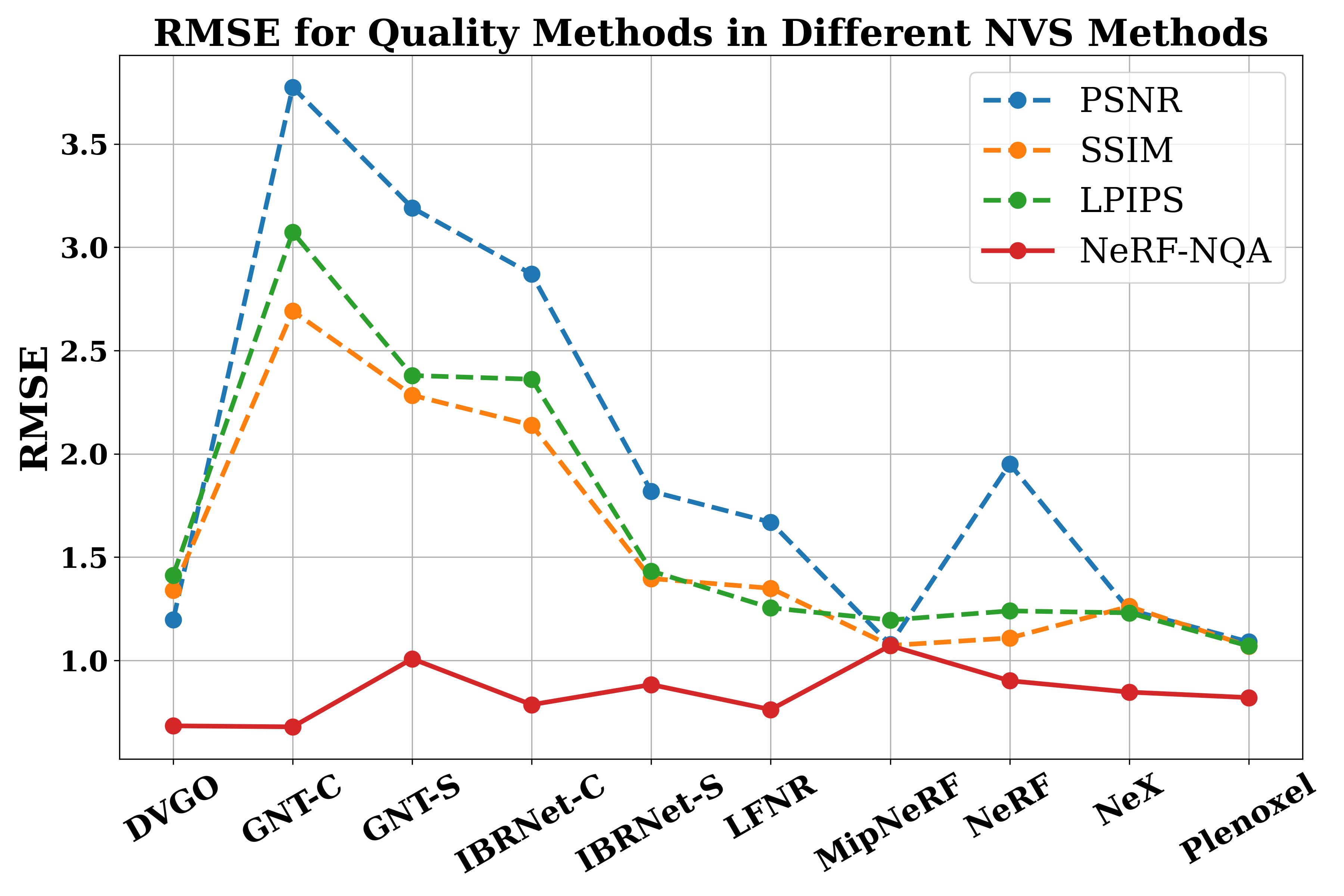}
}\\
\vspace{-0.8em}
\subfloat[]{
    \includegraphics[width=0.85\columnwidth]{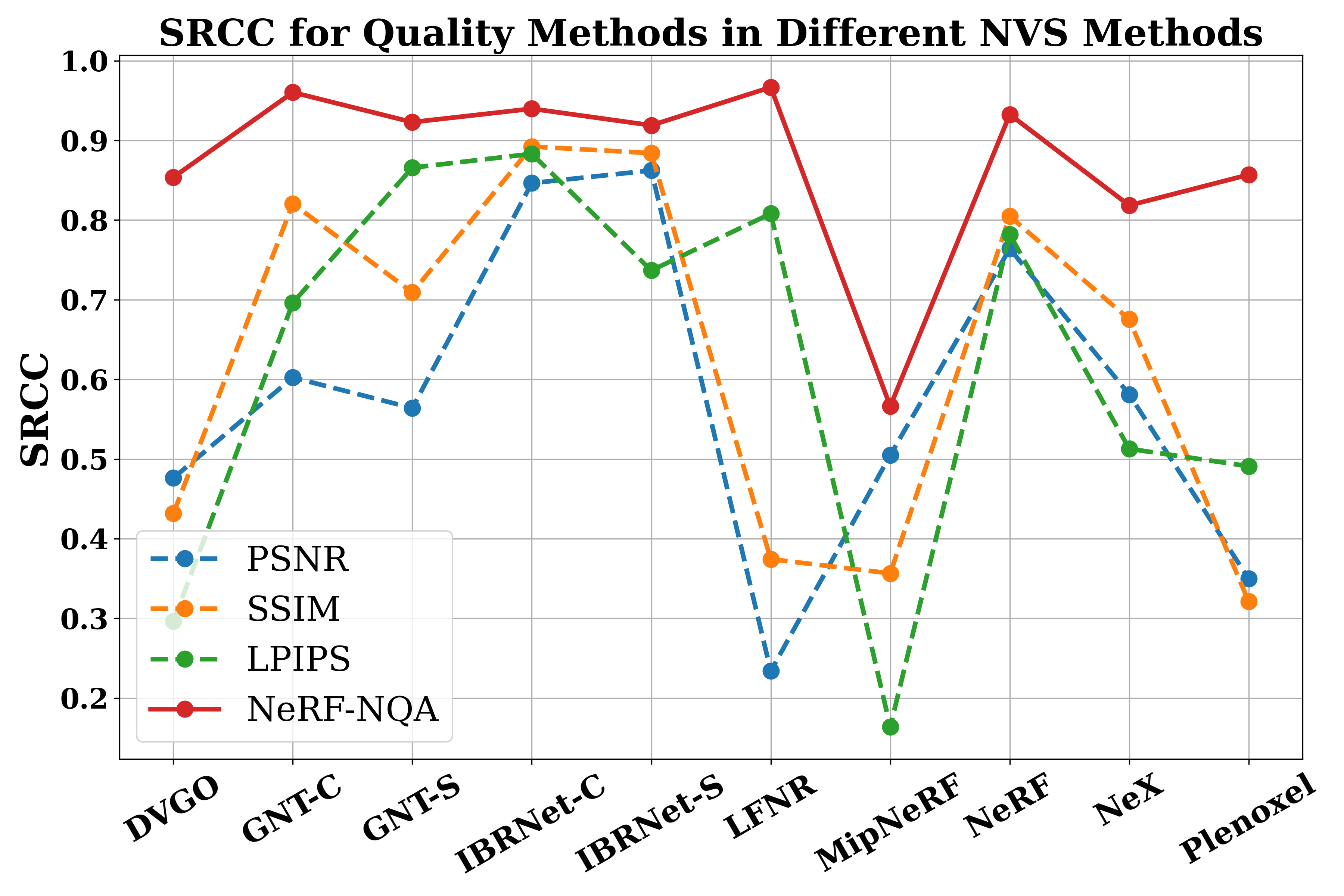}
}
\vspace{-0.8em}
\caption{\textbf{Quantitative Evaluation of PSNR, SSIM, LPIPS and NeRF-NQA Across Various NVS Methods}: (a) Line chart illustrating the RMSE $\downarrow$ performance for each NVS method; (b) Line chart depicting the SRCC $\uparrow$ values in relation to different NVS methods.}
\label{fig:line_charts}
\vspace{-1.2em}
\end{figure}

\subsection{Evaluation on Different Scenes}
To analyze the efficacy of the evaluated quality assessment methods across different scenes, we present the scene-wise performance statistics in Table~\ref{tab:scene_results}. As depicted in the table, the evaluation results of NeRF-NQA are consistently superior than others across a diverse array of NVS scenes. Specifically, NeRF-NQA attains the lowest RMSE values in 11 out of 12 scenes and the highest SRCC values in 10 out of 12 scenes. For RMSE, NeRF-NQA exhibits substantial improvements on scenes such as Dinosaur, Vespa, Horns, and Toys with enhancements of 77.0\%, 56.5\%, 55.1\%, and 54.1\%, respectively, compared to the second-best methods. Similarly, in terms of SRCC, NeRF-NQA demonstrates remarkable advantages on scenes like Flower, Fortress, and Room, improving SRCC by 0.458, 0.152, and 0.367, respectively, against the second-best methods.

Figure~\ref{fig:bench_plots} presents scatter plots contrasting ground truth quality scores with estimations from widely-used NVS methods (i.e., PSNR, SSIM, and LPIPS) as well as the proposed NeRF-NQA, across the evaluated datasets. Each subplot includes a red line, symbolizing the ideal prediction trajectory where ground truth scores are equivalent to estimations. The closeness of data points to this red line serves as an indicator of the method's predictive accuracy. Upon scrutinizing the first row of Figure~\ref{fig:bench_plots} pertaining to the Fieldwork dataset, it becomes evident that conventional methods like PSNR, SSIM, and LPIPS tend to produce biased estimations, particularly overestimating quality scores in scenes such as Dinosaur, Naiad-Sta., and Vespa. In contrast, NeRF-NQA effectively mitigates such biases across all scenes. Further analysis of the second and third rows of Figure~\ref{fig:bench_plots} reveals that NeRF-NQA's estimations are remarkably more concentrated and closely aligned with the red line, representing the ideal prediction trajectory, compared to other methods. This underscores that NeRF-NQA not only estimates with reduced bias but also with lower variance and a significantly diminished presence of outliers across all NVS scenes.

Figure~\ref{fig:example_scenes} presents three illustrative NVS scenes generated by three different NVS methods to qualitatively demonstrate the efficacy of NeRF-NQA. Accompanying each scene are the ground truth Just-Noticeable Differences (JOD) scores, along with estimations from PSNR, SSIM, LPIPS, and NeRF-NQA. The results compellingly indicate that NeRF-NQA's estimations are in close alignment with the ground truth JOD scores. For example, in the Dinosaur scene, which is characterized by pronounced blur and artifacts, conventional methods such as PSNR, SSIM, and LPIPS significantly overestimate the JOD score. In contrast, NeRF-NQA's estimation stands at -8.6999, remarkably close to the ground truth score of -8.7655. A similar pattern is observed in the subsequent examples; while other methods either overestimate the JOD score in the Naiad-Sta. scene or underestimate it in the Horns scene, NeRF-NQA consistently produces estimations that closely approximate the ground truth JOD scores.

\subsection{Evaluations on Different NVS Methods}

Table~\ref{tab:nvs_method_results} lists the performance of the evaluated quality assessment methods in different NVS methods. As delineated in Table~\ref{tab:nvs_method_results}, NeRF-NQA consistently outperforms other quality assessment methods across a diverse range of NVS methods. Specifically, NeRF-NQA achieves the most best RMSE values in 9 of the 10 evaluated NVS methods and the highest SRCC values across all NVS methods. In the context of RMSE, NeRF-NQA manifests significant performance gains in methods such as GNT-C, GNT-S, IBRNet-C, and LFNR, registering improvements of 74.8\%, 55.8\%, 63.3\%, and 38.1\%, respectively, when compared to the second-best performing metrics. Likewise, with respect to SRCC, NeRF-NQA exhibits pronounced advantages in methods like DVGO, NeRF, and Plenoxel, enhancing SRCC values by 0.377, 0.128, and 0.165, respectively, relative to the next best-performing metrics.

For a more comprehensive understanding, the line charts presented in Figure \ref{fig:line_charts} visualize the performance of NeRF-NQA in terms of RMSE and SRCC metrics, juxtaposed with prevalent NVS methods such as PSNR, SSIM, and LPIPS across various NVS methods. Thus, the figure apparently show that NeRF-NQA excels among its competing methods for quality assessment across different NVS methods.

\textcolor{HighlightColor}{To rigorously evaluate the robustness of the proposed NeRF-NQA method in challenging scenarios, we have quantitatively assessed its performance across different scene types, as detailed in Table~\ref{tab:special_cases_results}. This evaluation specifically targets special scenes traditionally deemed difficult for quality assessment, including those with complex shapes and specular objects, while also incorporating standard scenes for a comprehensive comparison. The empirical results consistently demonstrate that NeRF-NQA significantly surpasses competing methods, with a notably higher margin of improvement in special cases. This enhanced performance, particularly marked in complex and specular scenarios as evidenced in the last column of the table, underscores the method's robustness and efficacy. Such outcomes are likely attributable to the capacity of pointwise module on mitigating viewpoint dependency, affirming its integral role in the method's success.}

\begin{table*}[htbp]

\caption{\textcolor{HighlightColor}{\textbf{Comparative analysis of quality assessment methods for special cases including complex-shaped and specular scenes along with normal cases.}, using SRCC and PLCC metrics. Complex-shaped scenes include objects such as plants and skeletal specimens, while specular surfaces encompass scenes with specular reflections and transparent objects. The penultimate column presents the rankings of NeRF-NQA. The final column delineates either the enhancement achieved by NeRF-NQA over the second-leading method (when NeRF-NQA is top-ranked) or the difference relative to the foremost method (if NeRF-NQA doesn't achieve the best score).}}

\label{tab:special_cases_results}

\scriptsize
\renewcommand{\arraystretch}{1.0}
\setlength{\tabcolsep}{2pt}
{\color{HighlightColor}\begin{tabular}{P{0.11\textwidth}|P{0.09\textwidth}|P{0.07\textwidth}P{0.06\textwidth}P{0.07\textwidth}P{0.07\textwidth}P{0.07\textwidth}P{0.08\textwidth}P{0.07\textwidth}P{0.08\textwidth}|P{0.04\textwidth}|P{0.09\textwidth}}
\hline
\hline
Special Case & Evaluation & PSNR & SSIM & LPIPS & BRISQUE & VMAF & VIIDEO & LFACon
 & \textbf{NeRF-NQA} & \textbf{Rank}& \textbf{Against Best Alt. Method}\\
\hline

\multirow{ 2 }{*}{Complex Shapes} & SRCC ↑ & $0.7093$ & $0.6424$ & $0.2842$ & $0.0936$ & $0.3027$ & $0.3996$ & $0.5076$ & $0.9450$ & 1 & +0.236 \\ \cline{2-2} \cline{11-12}
 & PLCC ↑ & $0.6916$ & $0.6135$ & $0.3628$ & $0.1085$ & $0.2931$ & $0.4282$ & $0.5974$ & $0.9674$ & 1 & +0.276 \\ \cline{1-12}
\multirow{ 2 }{*}{Specular Surfaces} & SRCC ↑ & $0.3302$ & $0.1847$ & $0.1616$ & $-0.2457$ & $0.1482$ & $-0.0094$ & $0.5702$ & $0.7330$ & 1 & +0.163 \\ \cline{2-2} \cline{11-12}
 & PLCC ↑ & $0.2473$ & $0.2459$ & $0.1438$ & $-0.2311$ & $0.1328$ & $0.0890$ & $0.6444$ & $0.7962$ & 1 & +0.152 \\ \cline{1-12}
\multirow{ 2 }{*}{Normal Cases} & SRCC ↑ & $0.7934$ & $0.8740$ & $0.7727$ & $-0.1422$ & $0.6615$ & $-0.0528$ & $0.5392$ & $0.9431$ & 1 & +0.069 \\ \cline{2-2} \cline{11-12}
 & PLCC ↑ & $0.7917$ & $0.8226$ & $0.7310$ & $-0.1164$ & $0.6793$ & $-0.0102$ & $0.5816$ & $0.9591$ & 1 & +0.137 \\ \cline{1-12}

\hline
\hline
\end{tabular}}
\end{table*}

\subsection{Cross Dataset Evaluation}
\textcolor{HighlightColor}{To substantiate the model's generalizability, cross-dataset evaluations were conducted, and the results are presented in Table~\ref{tab:cross_dataset_results}. This table delineates the model's efficacy when trained on two datasets and subsequently tested on the third, exemplified by the 'Fieldwork' column, which reflects results from training on the LLFF and Lab datasets. The results reveal that, for the case of dataset independence, the proposed method consistently surpasses competing approaches, affirming its superior generalization capabilities. Notably, when the Fieldwork dataset is the test set, NeRF-NQA achieves a significant 35.0\% improvement in RMSE over the second best method. Regarding the LLFF dataset, it exhibits a 13.2\% enhancement in RMSE, along with increments of 0.290 in SRCC and 0.257 in PLCC. Similarly, for the Lab dataset, NeRF-NQA secures substantial advancements, bringing a 20.9\% improvement in RMSE, a 0.133 increase in SRCC, and a 0.178 rise in PLCC, further evidencing its superior performance and generalization across diverse datasets.}

\begin{table*}[htb]

\caption{\textcolor{HighlightColor}{\textbf{Cross-dataset evaluation of various quality assessment methods}. The table presents results from cross-dataset testing, where each method is trained on two datasets and tested on the third. For instance, results in the 'Fieldwork' column are derived from models trained on the LLFF and Lab datasets. For each column, the best results are highlighted in bold, with the concluding row indicating the enhancement relative to the second-best result. The results of full-reference video quality assessment methods are ``-'' for LLFF because this dataset has no ground-truth videos.}}

\label{tab:cross_dataset_results}

\scriptsize
\renewcommand{\arraystretch}{1.0}
\setlength{\tabcolsep}{2pt}
{\color{HighlightColor}\begin{tabular}{P{0.06\textwidth}|P{0.125\textwidth}|P{0.058\textwidth}P{0.058\textwidth}P{0.058\textwidth}P{0.058\textwidth}|P{0.058\textwidth}P{0.058\textwidth}P{0.058\textwidth}P{0.058\textwidth}|P{0.058\textwidth}P{0.058\textwidth}P{0.058\textwidth}P{0.058\textwidth}}
\hline
\hline
&&\multicolumn{4}{c}{Fieldwork}&\multicolumn{4}{c}{LLFF}&\multicolumn{4}{c}{Lab}\\
\hline
Type & Method & RMSE  ↓ & SRCC ↑ & PLCC ↑& OR  ↓ & RMSE  ↓& SRCC ↑ & PLCC ↑ &OR  ↓ & RMSE  ↓ & SRCC ↑ & PLCC ↑ &OR  ↓ \\
\hline

\multirow{14}{*}{FR-IQA}&PSNR & $3.7940$ & $0.8278$ & $0.8323$ & $0.0056$ & $2.6261$ & $0.1698$ & $0.1813$ & $0.0000$ & $1.2514$ & $0.6046$ & $0.5452$ & $0.0000$\\
&SSIM & $3.3938$ & $0.8317$ & $0.9026$ & $0.0556$ & $1.8391$ & $0.2192$ & $0.2161$ & $0.0750$ & $1.1192$ & $0.5671$ & $0.3983$ & $0.1814$\\
&MS-SSIM & $3.5113$ & $0.8255$ & $0.8936$ & $0.0711$ & $1.8038$ & $0.2125$ & $0.2145$ & $0.0737$ & $1.1925$ & $0.5637$ & $0.3433$ & $0.1714$\\
&IW-SSIM & $3.4253$ & $0.8513$ & $0.9075$ & $0.0411$ & $1.6957$ & $0.2348$ & $0.2262$ & $0.0712$ & $1.6057$ & $0.5450$ & $0.3202$ & $0.1714$\\
&VIF & $3.8471$ & $0.8399$ & $0.8818$ & $\pmb{ 0.0000 }$ & $2.6166$ & $0.0434$ & $0.0649$ & $0.0000$ & $1.6289$ & $0.5328$ & $0.3665$ & $0.0000$\\
&FSIM & $3.6660$ & $0.8509$ & $0.9121$ & $0.0422$ & $1.6258$ & $0.2887$ & $0.2626$ & $0.0750$ & $1.2786$ & $0.5700$ & $0.3788$ & $0.1414$\\
&GMSD & $3.7750$ & $0.8462$ & $0.8751$ & $0.0067$ & $2.1477$ & $0.2438$ & $0.2587$ & $0.0187$ & $1.1446$ & $0.5676$ & $0.4185$ & $0.0000$\\
&VSI & $4.1231$ & $0.5140$ & $0.6343$ & $0.0256$ & $1.2846$ & $0.2352$ & $0.2669$ & $0.0400$ & $1.5947$ & $0.3535$ & $0.2868$ & $0.0157$\\
&DSS & $3.3479$ & $0.8799$ & $0.8647$ & $0.0000$ & $1.9236$ & $0.3962$ & $0.3910$ & $0.0250$ & $1.1607$ & $0.5827$ & $0.4799$ & $0.0000$\\
&HaarPSI & $3.5647$ & $0.8790$ & $0.8983$ & $0.0011$ & $2.0701$ & $0.2317$ & $0.2426$ & $0.0213$ & $1.8664$ & $0.5426$ & $0.3926$ & $0.0000$\\
&MDSI & $3.6792$ & $0.8614$ & $0.8811$ & $0.0122$ & $2.1075$ & $0.2719$ & $0.2780$ & $0.0250$ & $1.3003$ & $0.5591$ & $0.4668$ & $0.0000$\\
&LPIPS & $3.5493$ & $0.8173$ & $0.8806$ & $0.0189$ & $2.0290$ & $0.1255$ & $0.1944$ & $0.0488$ & $1.3633$ & $0.3351$ & $0.3333$ & $0.0271$\\
&PieAPP & $3.5930$ & $0.8626$ & $0.8707$ & $0.0300$ & $3.4550$ & $0.1613$ & $0.2252$ & $0.0250$ & $0.8802$ & $0.7266$ & $0.6461$ & $0.0429$\\
&DISTS & $3.4913$ & $0.8554$ & $0.9112$ & $0.0622$ & $1.4191$ & $0.3393$ & $0.3507$ & $0.0275$ & $1.0460$ & $0.3167$ & $0.3242$ & $0.0157$\\
\hline
\multirow{3}{*}{NR-IQA}&BRISQUE & $4.1241$ & $-0.1489$ & $-0.1253$ & $0.0022$ & $1.8850$ & $-0.0759$ & $-0.0781$ & $0.0000$ & $2.6836$ & $-0.3995$ & $-0.4152$ & $0.0000$\\
&NIQE & $4.2406$ & $0.0296$ & $0.0078$ & $0.0033$ & $1.4894$ & $0.0649$ & $0.0379$ & $0.0000$ & $1.6843$ & $0.3635$ & $0.2811$ & $0.0000$\\
&CLIP-IQA & $4.5593$ & $-0.4992$ & $-0.5361$ & $0.0000$ & $1.5487$ & $-0.0831$ & $-0.0870$ & $0.0000$ & $1.5275$ & $-0.4851$ & $-0.3970$ & $0.0000$\\
\hline
\multirow{4}{*}{VQA}&STRRED & $1.8680$ & $0.9002$ & $0.8778$ & $0.0911$ & $-$ & $-$ & $-$ & $-$ & $1.3699$ & $0.5700$ & $0.4939$ & $0.0400$\\
&VMAF & $4.1449$ & $0.8128$ & $0.8283$ & $0.0011$ & $-$ & $-$ & $-$ & $-$ & $1.7342$ & $-0.2734$ & $-0.2234$ & $0.0057$\\
&FovVideoVDP & $4.2092$ & $0.6552$ & $0.7067$ & $0.0200$ & $-$ & $-$ & $-$ & $-$ & $2.0782$ & $-0.4587$ & $-0.4000$ & $0.0000$\\
&VIIDEO & $4.0178$ & $0.2579$ & $0.3029$ & $0.0111$ & $1.8884$ & $0.3590$ & $0.3817$ & $0.0000$ & $1.4648$ & $0.3241$ & $0.2732$ & $0.0100$\\
\hline
\multirow{2}{*}{LFIQA}&ALAS-DADS & $2.6011$ & $0.5655$ & $0.6553$ & $0.0700$ & $1.1131$ & $0.4886$ & $0.4914$ & $0.0000$ & $1.0946$ & $0.3126$ & $0.2366$ & $0.0000$\\
&LFACon & $2.4677$ & $0.5662$ & $0.6778$ & $0.2800$ & $1.1503$ & $0.3700$ & $0.4374$ & $0.0537$ & $0.8275$ & $0.5708$ & $0.6629$ & $0.0900$\\
\hline

&\textbf{NeRF-NQA}  & $\pmb{ 1.2138 }$ & $\pmb{ 0.9125 }$ & $\pmb{ 0.9457 }$ & $0.0444 $ & $\pmb{ 0.9666 }$ & $\pmb{ 0.7785 }$ & $\pmb{ 0.7487 }$ & $\pmb{ 0.0000 }$ & $\pmb{ 0.6549 }$ & $\pmb{ 0.8600 }$ & $\pmb{ 0.8408 }$ & $\pmb{ 0.0000 }$\\
&\textbf{Boost v.s. 2nd Best}  & $ \pmb{ +35.0\% } $ & $ \pmb{ +0.012 } $ & $ \pmb{ +0.034 } $ & $ \pmb{ -0.044 } $ & $ \pmb{ +13.2\% } $ & $ \pmb{ +0.290 } $ & $ \pmb{ +0.257 } $ & $ \pmb{ - } $ & $ \pmb{ +20.9\% } $ & $ \pmb{ +0.133 } $ & $ \pmb{ +0.178 } $ & $ \pmb{ - } $\\

\hline
\hline
\end{tabular}}
\end{table*}

\section{Limitation}

A notable limitation of the proposed method is its reliance on the sparse points generated by COLMAP~\cite{schonberger2016structure} within the pointwise module. Despite this dependency, the empirical evidence presented in Table~\ref{tab:special_cases_results} suggests that this reliance does not markedly diminish the model's performance, even in scenarios traditionally challenging for sparse point generation, such as scenes with complex shapes and specular surfaces. The sustained performance in these conditions indicates a degree of resilience to the noise inherent in COLMAP-generated sparse points. Future work will aim to address and potentially mitigate this dependency on COLMAP to further enhance the robustness and applicability of the proposed method. Due to time constraints, this research primarily focused on front-facing scenarios. As a result, perceptual scores for 360-degree scenes were not collected, and the methods related to 360-degree scenes were not tested. Additionally, the research did not encompass some of the latest advancements in NVS methods, including Instant-NGP~\cite{muller2022instant}, TensorRF~\cite{chen2022tensorf}, and 3D Gaussian Splatting~\cite{kerbl20233d}.
Future work will aim to address these gaps by incorporating evaluations on 360-degree scenes and integrating a wider array of NVS methods to provide a more exhaustive analysis of the proposed approach.

\section{Conclusion}
In this paper, we introduce NeRF-NQA, an innovate quality assessment method to evaluate the quality of NVS-generated scenes without the dependency on reference views, addressing the prevalent challenges on scarce reference availability in NVS scenarios. NeRF-NQA adopts a joint quality assessment strategy, integrating both viewwise and pointwise assessment methodologies to
facilitate a holistic evaluation of both the spatial fidelity and the intricate angular quality of the synthesized views.
Empirical results underscore the pronounced superiority of NeRF-NQA in gauging the quality of NVS-generated views, outperforming extant quality assessment techniques for images, videos, and light fields. These findings accentuate the efficacy and robustness of NeRF-NQA as a pivotal instrument for discerning the perceptual quality of NVS-generated scenes.

\acknowledgments{%
This work was supported in part by Beijing Natural Science Foundation (No. 4222003), National Natural Science Foundation of China (No. 62177001), and Shandong Provincial Natural Science Foundation (No. 2022HWYQ-040).
}

\bibliographystyle{abbrv-doi-hyperref}

\bibliography{template}

\newpage

\appendix 







\end{document}